\documentclass[10pt,twocolumn,letterpaper]{article}

\usepackage{cvpr}
\usepackage{times}
\usepackage{epsfig}
\usepackage{graphicx}
\usepackage{tabularx}
\usepackage{amsmath}
\usepackage{amssymb}
\usepackage{bm}
\usepackage{bbm}
\usepackage{algorithm}
\usepackage{algorithmic}
\usepackage{booktabs}
\usepackage{multirow}
\usepackage{tabularx}

\usepackage{array}
\newcolumntype{?}{!{\vrule width 1pt}}
\newcolumntype{C}[1]{>{\centering\arraybackslash\hspace{0pt}}p{#1}}

\usepackage{color}

\newif\ifdraft
\draftfalse

\newcommand{\comment}[1]{}

\ifdraft
 \newcommand{\ms}[1]{{\color{green}{#1}}}
 
 \newcommand{\ar}[1]{{\color{blue}{#1}}}
 \newcommand{\pf}[1]{{\color{red}{#1}}}
 \newcommand{\PF}[1]{{\color{red}{\bf #1}}}
 \newcommand{\MS}[1]{{\color{green}{\bf #1}}}
  \newcommand{\AR}[1]{{\color{blue}{\bf #1}}}
\else
 \newcommand{\ms}[1]{ #1 }
 
 \newcommand{\ar}[1]{ #1 }
 \newcommand{\pf}[1]{ #1 }
 \newcommand{\AR}[1]{}
 \newcommand{\PF}[1]{}
 \newcommand{\MS}[1]{}
\fi

\newcommand{\bx}[0]{\mathbf{x}}

\newcommand{\bX}{\mathbf{X}}
\newcommand{\bff}{\mathbf{f}}

\newcommand{\mdsz}{0.25in}

\newcommand{\fig}[1]{Fig.~\ref{fig:#1}}
\newcommand{\sect}[1]{Section~\ref{sec:#1}}

\newcommand{\tbl}[1]{Table~\ref{tbl:#1}}

\newcommand{\eqt}[1]{Eq.~\ref{eq:#1}}
\newcommand{\eqts}[2]{Eqs.~\ref{eq:#1},~\ref{eq:#2}}
\newcommand{\eqtss}[3]{Eqs.~\ref{eq:#1},~\ref{eq:#2},~\ref{eq:#3}}
\newcommand{\eqtsss}[4]{Eqs.~\ref{eq:#1},~\ref{eq:#2},~\ref{eq:#3},~\ref{eq:#4}}

\newcommand{\mysum}[0]{\mathop{\sum}\limits}
\newcommand{\norm}[1]{\left\lVert#1\right\rVert}
\newcolumntype{M}[1]{>{\centering\arraybackslash}m{#1}}
\newcolumntype{R}[1]{>{\raggedleft\arraybackslash}m{#1}}
\newcolumntype{P}[1]{>{\centering\arraybackslash}p{#1}}

\newcommand{\hl}[1]{\textbf{#1}}

\usepackage[pagebackref=true,breaklinks=true,letterpaper=true,colorlinks,bookmarks=false]{hyperref}

\cvprfinalcopy 

\def\cvprPaperID{523} 

\ifcvprfinal\fi
\begin{document}

\title{Beyond Sharing Weights for Deep Domain Adaptation}

\author{Artem Rozantsev \qquad Mathieu Salzmann \qquad Pascal Fua \\
  Computer Vision Laboratory, \'{E}cole Polytechnique F\'{e}d\'{e}rale de Lausanne\\
  Lausanne, Switzerland\\
  {\tt\small \{firstname.lastname\}@epfl.ch}
}

\maketitle

\begin{abstract}

The performance of a classifier trained on data coming from a specific domain typically degrades when applied to a related but different one. While annotating many samples from the new domain would address this issue, it is often too expensive or impractical. Domain Adaptation has therefore emerged as a solution to this problem; It leverages annotated data from a source domain, in which it is abundant, to train a classifier  to operate  in a  target
  domain, in which it is either sparse or even lacking altogether. In this context, the recent trend consists of learning deep architectures whose weights are shared for both domains, which  essentially amounts  to
learning domain invariant features.

Here, we  show that it is more effective to
explicitly  model the  shift from  one domain  to the  other.  To  this end,  we
introduce a two-stream architecture, where one operates in the source domain
and  the other  in the  target  domain. In  contrast to  other approaches,  the
weights in corresponding layers are related  but {\it not shared}. We demonstrate that this both yields higher
accuracy  than  state-of-the-art  methods  on  several  object  recognition  and
detection tasks  and consistently  outperforms networks  with shared  weights in
both supervised and unsupervised settings.

\comment{Deep Neural Networks have demonstrated  outstanding performance in many Computer
Vision tasks  but typically require  large amounts  of labeled training  data to
achieve  it.  This  is  a serious  limitation  when such  data  is difficult  to
obtain. In  traditional Machine  Learning, Domain Adaptation  is an  approach to
overcoming this problem by leveraging annotated data from a {\it source domain},
in which  it is  abundant, to  train a classifier  to operate  in a  {\it target
  domain}, in which labeled data is either sparse or even lacking altogether. In
the Deep Learning case, most existing methods use the same architecture with the
same  weights for  both source  and target  data, which  essentially amounts  to
learning domain invariant features.  Here, we  show that it is more effective to
explicitly  model the  shift from  one domain  to the  other.  To  this end,  we
introduce a two-stream architecture, where one operates in the source domain
and  the other  in the  target  domain.  In  contrast to  other approaches,  the
weights in corresponding layers are related  but {\it not shared} to account for
differences between the two domains. We demonstrate that this both yields higher
accuracy  than  state-of-the-art  methods  on  several  object  recognition  and
detection tasks  and consistently  outperforms networks  with shared  weights in
both supervised and unsupervised settings.}

\end{abstract}


\section{Introduction}

  A classifier  trained using samples from  a specific domain usually  needs to be
re-trained  to perform  well in  a related  but different  one.  Since  this may
require  much manual  annotation to  create enough  training data,  it is  often
impractical.  \ms{With  the advent of Deep  Networks~\cite{Hinton06a,LeCun98b}, this
problem  has become  particularly acute  due to  their requirements  for massive
amounts of training data.
}

Domain  Adaptation~\cite{Jiang08} and  Transfer Learning~\cite{Pan10}  have long
been used to overcome this difficulty by  making it possible to exploit what has
been learned in one source domain,  for which enough training data is available,
to  effectively train  classifiers in  a target  domain, where  only very  small
amounts of  additional annotations,  or even none,  can be  acquired.  Recently,
Domain Adaptation  has been investigated  in the  context of Deep  Learning with
promising     results~\cite{Girshick13,Oquab14,Tzeng14,Long15b,Ganin15,Tzeng15}.
These methods, however, use the same deep architecture with the same weights for
both source and  target domains. In other words, they  attempt to learn features
that are invariant to the domain shift.

\begin{figure}[!t]
	\centering
	\includegraphics[width = \linewidth]{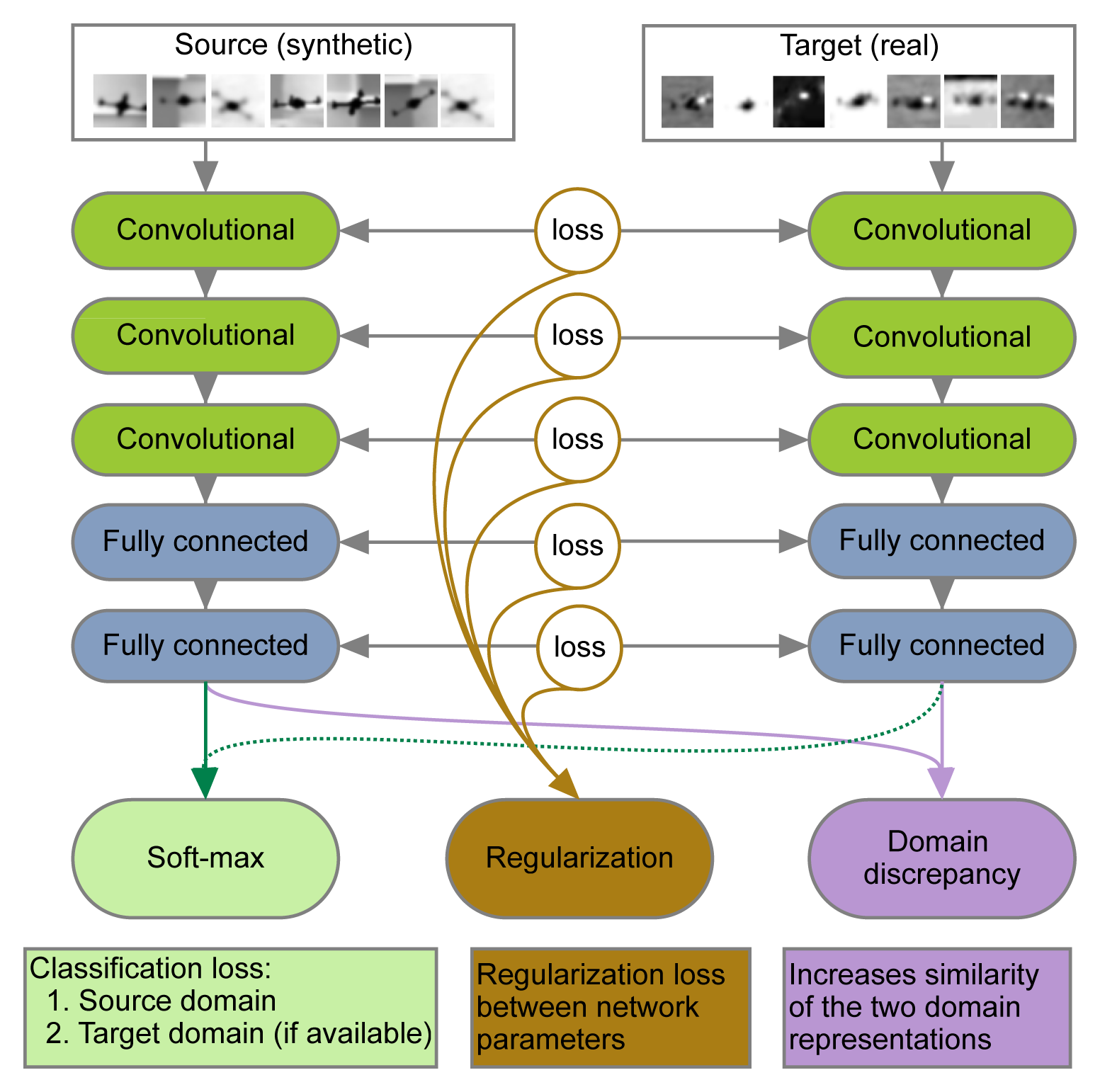}
	\caption{{\bf Our  two-stream architecture.} One stream  operates on
		the source data and the other on the target data. Their weights are
		{\it not} shared.  Instead, we introduce loss functions that prevent
		corresponding weights from being too different from each other.}
	\label{fig:streams}
\end{figure}

In  this paper,  we  show that  imposing feature  invariance  is detrimental  to
discriminative  power. To  this end,  we introduce  the two-stream  architecture
depicted by~\fig{streams}.   One stream  operates on the  source domain  and the
other on the target one.  This makes  it possible {\it not} to share the weights
in some of the layers. Instead, we introduce a loss function that is lowest when
they are  linear transformations  of each  other. \ms{Furthermore, we introduce  a
criterion to automatically determine which layers  should share  their weights and  which ones should
not.}  In short,  our approach  explicitly models  the domain  shift by  learning
features adapted to  each domain, but not fully independent,  to account for the
fact that both domains depict the same underlying problem.

We  demonstrate  that  our  approach is  more  effective  than  state-of-the-art
weight-sharing  schemes  on  \ms{standard  Domain Adaptation benchmarks for  image  recognition}.  We also  show that it is well suited  to \ms{leveraging} synthetic data
to increase the performance of a classifier  on real images.  Given that this is
one of the easiest ways to provide  the large amounts of training data that Deep
\ms{Networks}  require,  this  scenario  has  become popular.   Here,  we  treat  the
synthetic  images as  forming the source  domain  and the  real images  the target one. We then
\ms{make use of} our  two-stream architecture to  learn an effective \ms{model} for the
real  data  even  though  we  have only \ms{few annotations for it}.   We
demonstrate \ms{the effectiveness of our approach at leveraging synthetic data}  for both detection  of Unmanned Aerial  Vehicles (UAVs)  and facial
pose estimation.  The  first application involves classification  and the second
regression, \ms{and}  they both benefit from  using synthetic data. We  outperform the
state-of-the-art  methods in  all these  cases, and  our experiments support our
contention that specializing the network weights outperforms sharing them.


\section{Related Work}
\label{sec:related}

In many practical  applications, classifiers and regressors may  have to operate
on various kinds of related but visually different image data. The differences are
often large enough for  an algorithm that has been trained on  one kind of images
to perform  poorly on another. Therefore,  new training data has  to be acquired
and  annotated   to  re-train  it.    Since  this  is  typically   expensive  and
time-consuming,  there  has  long  been  a push  to  develop  Domain  Adaptation
techniques that allow re-training with minimal amount of new data or even none.
Here, we briefly review some recent trends, with a focus  on Deep Learning based
methods, which are the most related to our work.

A natural approach to Domain Adaptation is to modify a classifier trained on the
source data using the available labeled target data. This was done, for example,
using SVM~\cite{Duan09,Bergamo10},     Boosted     Decision
Trees~\cite{Becker13d} and other classifiers~\cite{Daume06}. In the
context  of  Deep  Learning,  fine-tuning~\cite{Girshick13,Oquab14}  essentially
follows this pattern.  In practice, however,  when only a small amount of labeled target data is
available, this often results in overfitting.

Another approach is to learn a metric  between the source and target data, which
can also be interpreted  as a linear cross-domain transformation~\cite{Saenko10}
or a non-linear one~\cite{Kulis11}. Instead  of working on the samples directly,
several   methods   involve   representing   each   domain   as   one   separate
subspace~\cite{Gopalan11,Gong12b,Fernando13,Caseiro15}. A  transformation can then  
be learned to  align them~\cite{Fernando13}.   Alternatively,   one  can  interpolate
between  the  source   and  target  subspaces~\cite{Gopalan11,Gong12b,Caseiro15}.
In~\cite{Chopra13}, this  interpolation idea  was extended  to Deep  Learning by
training  multiple  unsupervised  networks  with increasing  amounts  of  target
data. The  final representation of  a sample  was obtained by  concatenating all
intermediate ones.  It is unclear, however, why this concatenation should be meaningful
to classify a target sample.

Another way to  handle the domain shift is  to explicitly try making  the source and
target  data distributions  similar.  While  many metrics  have been  proposed to
quantify the similarity  between two distributions, the most widely  used in the Domain Adaptation context is 
the Maximum  Mean  Discrepancy  (MMD)~\cite{Gretton07}. The MMD has been used to re-weight~\cite{Huang06,Gretton09} or
select~\cite{Gong13} source samples such that the resulting distribution becomes
as  similar as  possible  to the  target  one.   An alternative  is  to learn  a
transformation of  the data,  typically both  source and  target, such  that the
resulting    distributions    are    as    similar   as    possible    in    MMD
terms~\cite{Pan09,Muandet13,Baktash13}.  In~\cite{Ghifary14},  the MMD  was
used within a shallow neural  network architecture.  However, this method relied
on SURF features~\cite{Bay08} as initial image representation and thus only
achieved limited accuracy.

Recently,  using  Deep  Networks  to  learn features  has  proven  effective  at
increasing the  accuracy of Domain Adaptation  methods.  In~\cite{Donahue13a}, it
was  shown  that using  DeCAF  features  instead  of hand-crafted  ones 
mitigates  the  domain  shift  effects  even  without  performing  any  kind  of
adaptation.   However, performing  adaptation within  a Deep  Learning framework  was shown to boost accuracy even further~\cite{Chopra05,Tzeng14,Long15b,Ganin15,Tzeng15,Ghifary16,Bousmalis16}.   For example, in~\cite{Chopra05}, a Siamese architecture was  introduced to minimize the distance between pairs of source and target samples, which requires training labels available in the \emph{target} domain thus making the method unsuitable for unsupervised Domain Adaptation. The MMD has also been used to relate the source and  target data representations learned by Deep Networks~\cite{Tzeng14,Long15b} thus making  it possible to avoid working on individual samples. \cite{Ganin15,Tzeng15} introduced a loss term that  encodes an  additional classifier predicting  from which  domain each sample comes.  This was motivated by  the fact  that, if the learned features  are domain-invariant, such a classifier should  exhibit very poor performance.

All these Deep Learning approaches rely on the same architecture with the same weights for both the  source and target domains.   In essence, they attempt  to reduce the
impact  of  the  domain  shift  by  learning  domain-invariant  features.   In
practice,  however,  domain  invariance  might very  well  be  detrimental  to
discriminative  power.   As  discussed  in   the  introduction,  this  is  the
hypothesis we  set out to  test in this work  by introducing an  approach that explicitly
models the domain shift instead of attempting  to enforce invariance to it. \comment{We will see} We show in the  results  section that  this  yields  a  significant accuracy  boost  over networks with shared weights.

\comment{
Here, rather than trying to overcome the domain shift, as done by Deep Domain
Confusion method (DDC)~\cite{Tzeng14}, we propose to explicitly account for it.
DDC method is closest in spirit to ours, as we also use the MMD loss between 
learned source and target representations and rely on Deep Network for feature
extraction. We therefore use DDC as a main baseline for our approach, as the core
difference is that our network does not rely on shared weights for source and 
target domains, while the one in DDC does. As evidenced  by  our experiments, our 
approach yields a significant boost in  accuracy compared to the networks with 
shared weights.
}

\comment{ Here, rather  than trying to overcome the domain  shift, we propose to
  explicitly  account  for  it.   We therefore  introduce  the  two-stream  deep
  architecture  depicted  by  Fig.~\ref{fig:streams}.  The weights  of  the  two
  streams are  not shared. They  can therefore be  different but we  introduce a
  loss that prevents  them from being too different from  those of corresponding
  layers.   We  also  use  the  MMD   between  the  learned  source  and  target
  representations.   This  combination  lets  us encode  the  fact  that,  while
  different,  the  two   domains  are  still  related.   As   evidenced  by  our
  experiments, our approach  yields a significant boost in  accuracy compared to
  the networks with shared weights.}


\section{Our Approach}
\label{sec:method}

\comment{In this section,  we introduce our Deep Learning approach  to Domain Adaptation.
At its heart lies  the idea that, } The core idea of our method is that, for \ms{a Deep Network} to  adapt to different domains,
the weights   should   be   related,  yet   different   for each of the two
domains. \ms{As shown empirically,} this constitutes a major \comment{difference between our  approach and the}\ms{advantage of our method over the} competing ones discussed in  Section~\ref{sec:related}. To implement this idea, we therefore introduce
a  two-stream architecture,  such as  the one  depicted by  \fig{streams}. The
first stream  operates on the  source data, the second  on the target  one, and
they are  trained jointly.  While we  allow the weights of  the corresponding
layers to differ between  the two streams, we prevent them from  being too far from
each other.\comment{ by introducing  appropriately designed
loss functions.}   Additionally we use  the MMD between  the learned source  and target
representations.   This  combination  lets  us encode  the  fact  that,  while
different, the two domains are related.

More formally, let $\bX^s = \{\bx^{s}_i\}_{i=1}^{N^s}$ and $\bX^t = \{\bx^{t}_i\}_{i=1}^{N^t}$ be the sets of training images from the source and target domains, respectively, with $Y^s = \{y^{s}_i\}$ and $Y^t = \{y^{t}_i\}$ being the corresponding labels. To handle unsupervised target data as well, we assume, without loss of generality, that the target samples are ordered, such that only the first $N^t_l$ ones have valid labels, where $N^t_l = 0$ in the unsupervised scenario. Furthermore, let $\theta_j^s$ and $\theta_j^t$ denote the parameters, \comment{i.e.,} that is, the weights and biases, of the $j^{\rm th}$ layer of the source and target streams, respectively. We train the network by \comment{formulate learning as the problem of }minimizing a loss function of the form
%
\begingroup\makeatletter\def\f@size{9}\check@mathfonts
\def\maketag@@@#1{\hbox{\m@th\normalfont#1}}%
\begin{align}
\small
L(\theta^s,\theta^t | \bX^{s}, Y^{s},\bX^{t}, Y^{t}) & = L_{s} + L_{t} + L_{w} + L_{MMD}, \label{eq:overall_loss}\\
L_{s} & = \frac{1}{N^s}\sum_{i=1}^{N^s}c(\theta^s | \bx^{s}_i,y^{s}_i), \label{eq:cost_rfs}\\
L_{t} & = \frac{1}{N^t_l}\sum_{i=1}^{N^t_l}c(\theta^t | \bx^{t}_i,y^{t}_i), \label{eq:cost_r}\\
L_{w} & = \lambda_w \sum_{j \in \Omega} r_w(\theta_j^s,\theta_j^t), \label{eq:cost_rf}\\
L_{MMD} & = \lambda_u r_u(\theta^s,\theta^t | \bX^{s},\bX^{t}), \label{eq:cost_rfssm} \;,
\end{align}\endgroup
where $c(\theta^{\cdot}|\bx^{\cdot}_i,y^{\cdot}_i)$ is a standard classification
loss, such as  the logistic loss or the hinge  loss. $r_w(\cdot)$
and  $r_u(\cdot)$  are  the   weight  and  unsupervised  regularizers  discussed
below. The first one represents the loss between corresponding layers of the two
streams. The second encodes the MMD  measure and favors similar distributions of
\ms{the source and target  data representations.} These regularizers are  weighted by
coefficients $\lambda_w$  and $\lambda_u$,  respectively. In practice,  we found
our approach to  be robust to the  specific values of these  coefficients and we
set them to $1$  in all our experiments. $\Omega$ denotes the  set of indices of
the layers whose  parameters are not shared. This set  is problem-dependent and,
in  practice,  can  be  obtained  by comparing  the  MMD  values  for  different
configurations, as demonstrated in our experiments.

\subsection{Weight Regularizer}
\label{sec:weightRegularizer}

While our goal is to go beyond  sharing the layer weights, we still believe that
corresponding weights in the two streams should be related. This \comment{would,
  on one hand, }models the fact that  the source and target domains are related,
and  prevents overfitting  in  the target  stream, when  only  very few  labeled
samples are available. Our  weight regularizer $r_w(\cdot)$ therefore represents
the distance  between the source  and target weights  in a particular  layer. In principle, we
could take  it to directly act on the difference of those weights. 
This, however, would not truly attempt to model the domain shift, for instance to account for different means and ranges of values in the two types of data. 
To better model the shift and introduce more flexibility in our model, we therefore propose not to penalize linear transformations between the source and target weights. We then write our regularizer either by relying on the $L_2$ norm as
\begin{equation}
r_w(\theta_j^s,\theta_j^t)  = \norm{a_j\theta_j^s+b_j-\theta_j^t}_2^2\;,
\label{eq:L2_loss_psi}
\end{equation}
\ms{or in an exponential form as}
\begin{equation}
r_w(\theta_j^s,\theta_j^t)                                                     =
\exp{\left(\|a_j\theta_j^s+b_j-\theta_j^t\|^2\right)}-1 \; .
\label{eq:exp_loss_psi}
\end{equation}
\ms{In both cases,} $a_j$ and $b_j$ are scalar parameters that are different for each layer $j
\in  \Omega$  and  learned  at  training  time  along  with  all  other  network
parameters.   While  simple,  this   parameterization  can  account, for example, for  global
illumination changes in  the first layer of  the network. As shown  in the
results section, we found empirically that the exponential version gives better results.

We have tried replacing the simple linear transformation of Eqs.~\ref{eq:L2_loss_psi} and~\ref{eq:exp_loss_psi} by
more sophisticated  ones, such as quadratic  or piecewise linear ones.  This, however,
did not yield any performance improvement.

\subsection{Unsupervised Regularizer}

In addition to regularizing the weights of corresponding layers in the two streams, we also aim at learning a final representation, that is, the features before the classifier layer, that is domain invariant. To this end, we introduce a regularizer $r_u(\cdot)$ designed to minimize the distance between the distributions of the source and target representations. Following the popular trend in Domain Adaptation~\cite{Long13a,Tzeng14}, we rely on the Maximum Mean Discrepancy (MMD)~\cite{Gretton07} to encode this distance.

As the  name suggests,  given two sets  of data, the  MMD measures  the distance
between the  mean of  the two sets  after mapping each  sample to  a Reproducing
Kernel   Hilbert   Space    (RKHS).   In   our   context,    let   $\bff^s_i   =
\bff^s_i(\theta^s,\bx^s_i)$ be the  feature representation at the  last layer of
the source stream, and $\bff^t_j   =
\bff^t_j(\theta^t,\bx^t_j)$ of the target stream. The $\text{MMD}^2$ between
the source and target domains can be expressed as
\begin{equation}
\small
\text{MMD}^2(\{\bff^s_i\},\{\bff^t_j\}) = \left\| \sum_{i=1}^{N^s} \frac{\phi(\bff^s_i)}{N^s} - \sum_{j=1}^{N^t} \frac{\phi(\bff^t_j)}{N^t} \right\|^2,
\label{eq:mmd_loss}
\end{equation}
\noindent
where $\phi(\cdot)$  denotes the mapping to  RKHS. In practice, this  mapping is
typically unknown.  Expanding \eqt{mmd_loss}  and using the kernel
trick to replace  inner products by kernel values lets  us rewrite the squared
MMD, and thus our regularizer \ms{as $r_u(\theta^s,\theta^t | \bX^s,\bX^t) = $}
  \begin{equation}
  \mysum_{i,i'} \frac{k(\bff^s_i,\bff^s_{i'})}{(N^s)^2} -  2 \mysum_{i,j}\frac{ k(\bff^s_i,\bff^t_j)}{N^sN^t} 
  + \mysum_{j,j'} \frac{k(\bff^t_j,\bff^t_{j'})}{(N^t)^2}\;,
   \label{eq:mmd_regularizer}
   \end{equation}
%
%
where the dependency on the  network parameters comes via the $\bff^{\cdot}_i$s,
and where $k(\cdot,\cdot)$ is a kernel function. In practice, we make use of the
standard  RBF kernel  $k(u,v)  =  \exp{(-\|u-v\|^2/\sigma)}$, with \ms{bandwidth $\sigma$.}
In  all our  experiments,  we found  our approach to be insensitive  to the choice of $\sigma$ and  we therefore set it
to $1$.

\subsection{Training}

To learn the model parameters\comment{ of our model}, we first pre-train \comment{ the weights of }the source stream using the source data only. We then simultaneously optimize the weights of both streams according to the loss of Eqs.~\ref{eq:cost_rfs}-\ref{eq:cost_rfssm} using both source and target data, with the target stream weights initialized from the pre-trained source weights.\comment{, and using the loss defined by Eqs.~\ref{eq:cost_rfs}-\ref{eq:cost_rfssm}} Note that this also requires initializing \comment{the parameters of }the linear transformation parameters of each layer, $a_j$ and $b_j$ for all $j \in \Omega$. We initialize these values to $a_j = 1$ and $b_j = 0$, thus encoding the identity transformation. All parameters are then learned jointly using backpropagation with the AdaDelta algorithm~\cite{Zeiler12}. Note that we rely on mini-batches, and thus in practice compute all the terms of our loss over these mini-batches rather than over the entire source and target datasets.
 
\comment{\begin{figure}[!h]
 	\centering
 	\includegraphics[width = 0.8\linewidth]{./fig/CNN_architecture_single}
 	\caption{We first pre-train the weights of the network using the data from the source domain.}
 	\label{fig:single_cnn}
\end{figure}}

Depending on the\comment{ domain adaptation} task, we use different network architectures, to provide a fair comparison with the baselines. For example, for the \emph{Office} benchmark, we adopt the AlexNet~\cite{Krizhevsky12} architecture, as was done in~\cite{Tzeng14}, and for digit classification we rely on the standard network structure of~\cite{LeCun98} for each stream. 



\section{Experimental Results}
\label{sec:results}


\begin{figure}[t!]
	\centering
	\begin{tabular}{cc}
		\toprule
		Synthetic & Real \\
		\begin{tabular}{cccccc}
			\hspace{-0.3cm}\multirow{2}{*}{\rotatebox{90}{\scriptsize{positives}}} &
			\hspace{-0.3cm}\includegraphics[width=\mdsz]{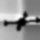} & 
			\hspace{-0.3cm}\includegraphics[width=\mdsz]{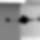} & 
			\hspace{-0.3cm}\includegraphics[width=\mdsz]{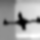} & 
			\hspace{-0.3cm}\includegraphics[width=\mdsz]{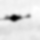} & 
			\hspace{-0.3cm}\includegraphics[width=\mdsz]{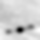} \\
			\vspace{0.1cm}
&			\hspace{-0.3cm}\includegraphics[width=\mdsz]{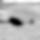} & 
			\hspace{-0.3cm}\includegraphics[width=\mdsz]{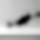} & 
			\hspace{-0.3cm}\includegraphics[width=\mdsz]{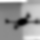} & 
			\hspace{-0.3cm}\includegraphics[width=\mdsz]{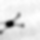} & 
			\hspace{-0.3cm}\includegraphics[width=\mdsz]{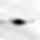} \\ 
			\hspace{-0.3cm}\multirow{2}{*}{\rotatebox{90}{\scriptsize{negatives}}} &
			\hspace{-0.3cm}\includegraphics[width=\mdsz]{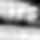} & 
			\hspace{-0.3cm}\includegraphics[width=\mdsz]{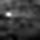} & 
			\hspace{-0.3cm}\includegraphics[width=\mdsz]{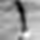} & 
			\hspace{-0.3cm}\includegraphics[width=\mdsz]{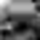} & 
			\hspace{-0.3cm}\includegraphics[width=\mdsz]{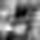} \\
&			\hspace{-0.3cm}\includegraphics[width=\mdsz]{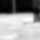} & 
			\hspace{-0.3cm}\includegraphics[width=\mdsz]{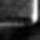} & 
			\hspace{-0.3cm}\includegraphics[width=\mdsz]{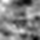} & 
			\hspace{-0.3cm}\includegraphics[width=\mdsz]{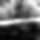} & 
			\hspace{-0.3cm}\includegraphics[width=\mdsz]{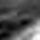} \\ 
		\end{tabular} &
		\hspace{-0.3cm}
		\begin{tabular}{ccccc}
			\hspace{-0.3cm}\includegraphics[width=\mdsz]{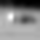} & 
			\hspace{-0.3cm}\includegraphics[width=\mdsz]{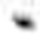} & 
			\hspace{-0.3cm}\includegraphics[width=\mdsz]{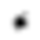} & 
			\hspace{-0.3cm}\includegraphics[width=\mdsz]{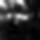} & 
			\hspace{-0.3cm}\includegraphics[width=\mdsz]{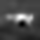} \\	
			\vspace{0.1cm}\hspace{-0.3cm}\includegraphics[width=\mdsz]{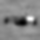} & 
			\hspace{-0.3cm}\includegraphics[width=\mdsz]{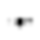} & 
			\hspace{-0.3cm}\includegraphics[width=\mdsz]{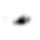} & 
			\hspace{-0.3cm}\includegraphics[width=\mdsz]{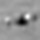} & 
			\hspace{-0.3cm}\includegraphics[width=\mdsz]{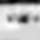} \\
			\hspace{-0.3cm}\includegraphics[width=\mdsz]{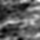} & 
			\hspace{-0.3cm}\includegraphics[width=\mdsz]{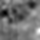} & 
			\hspace{-0.3cm}\includegraphics[width=\mdsz]{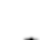} & 
			\hspace{-0.3cm}\includegraphics[width=\mdsz]{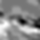} & 
			\hspace{-0.3cm}\includegraphics[width=\mdsz]{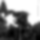} \\
			\hspace{-0.3cm}\includegraphics[width=\mdsz]{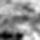} & 
			\hspace{-0.3cm}\includegraphics[width=\mdsz]{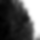} & 
			\hspace{-0.3cm}\includegraphics[width=\mdsz]{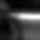} & 
			\hspace{-0.3cm}\includegraphics[width=\mdsz]{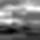} & 
			\hspace{-0.3cm}\includegraphics[width=\mdsz]{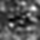} \\
		\end{tabular}
		\hspace{-0.3cm}
		\\
		\midrule
		\multicolumn{2}{c}{Test real data} \\
		\multicolumn{2}{c}{
		\begin{tabular}{cccccccccc}
			\hspace{-0.15cm} \raisebox{0.3cm}{\scriptsize{positives}} &
			\hspace{-0.3cm}\includegraphics[width=\mdsz]{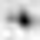} &
			\hspace{-0.3cm}\includegraphics[width=\mdsz]{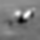} &
			\hspace{-0.3cm}\includegraphics[width=\mdsz]{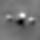} &
			\hspace{-0.3cm}\includegraphics[width=\mdsz]{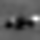} &
			\hspace{-0.3cm}\includegraphics[width=\mdsz]{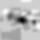} &
			\hspace{-0.3cm}\includegraphics[width=\mdsz]{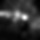} &
			\hspace{-0.3cm}\includegraphics[width=\mdsz]{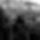} &
			\hspace{-0.3cm}\includegraphics[width=\mdsz]{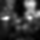} &
			\hspace{-0.3cm}\includegraphics[width=\mdsz]{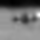} \\
			\hspace{-0.15cm}\raisebox{0.3cm}{\scriptsize{negatives}} & 
			\hspace{-0.3cm}\includegraphics[width=\mdsz]{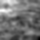} &
			\hspace{-0.3cm}\includegraphics[width=\mdsz]{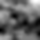} &
			\hspace{-0.3cm}\includegraphics[width=\mdsz]{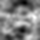} &
			\hspace{-0.3cm}\includegraphics[width=\mdsz]{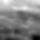} &
			\hspace{-0.3cm}\includegraphics[width=\mdsz]{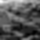} &
			\hspace{-0.3cm}\includegraphics[width=\mdsz]{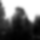} &
			\hspace{-0.3cm}\includegraphics[width=\mdsz]{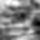} &
			\hspace{-0.3cm}\includegraphics[width=\mdsz]{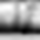} &
			\hspace{-0.3cm}\includegraphics[width=\mdsz]{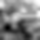} \\
		\end{tabular}	
		} \\
		\bottomrule
	\end{tabular}
	\caption{{\bf Our UAV dataset.} {\bf Top:} Synthetic and real training examples. {\bf Bottom:} Real samples from the test dataset.}
	\label{fig:UAV_db}

\vspace{-0.3cm}
\end{figure}

In  this section,  we demonstrate  the  potential of  our approach  in both  the
supervised  and unsupervised  scenarios using  different network  architectures.\comment{Since  our  motivating application  was  drone  detection, } We  first  thoroughly evaluate our method for \ar{the drone detection task.}\comment{this task.}  We then demonstrate that it generalizes well
to other \pf{classification problems} by testing it on the {\it Office} and {\it
  MNIST+USPS} datasets.  \comment{Due  to space the limitations, we only  report here our
results on the first and refer  the interested to the supplementary material for
results on the  second.  }Finally, to show that our  approach also generalizes to
regression problems, we apply it to estimating the position of facial landmarks.


\subsection{Leveraging Synthetic Data for Drone Detection}

\comment{As drones and  UAVs become ever more  numerous in our
skies, it  will become  increasingly important  for them to  see and  avoid colliding with each other. Unfortunately, }
\ar{Due to the lack of large publicly available datasets, UAV detection is a perfect example of a problem} where training  videos are scarce and do not cover a wide enough range of possible  shapes, poses, lighting  conditions, and backgrounds  against which \ar{drones} can be seen.   However, it is relatively easy to  generate large amounts of synthetic  examples, which  can be  used to  supplement a  small number  of real images and  increase detection accuracy~\cite{Rozantsev15b}.  We  show here that our approach  allows us to exploit these synthetic  images more effectively than  other state-of-the-art  Domain Adaptation  techniques. 

\subsubsection{Dataset and Evaluation Setup}

We used the  approach of~\cite{Rozantsev15b} to create a large  set of synthetic
examples. 
\fig{UAV_db} depicts sample images from the real and synthetic dataset that we used for training and testing.
In our experiments, we treat the synthetic images as source samples and the real images as target ones.

We report  results using  two versions of  this dataset, which  we refer  to as
\emph{UAV-200~\scriptsize{(small)}}  and  \emph{UAV-200~\scriptsize{(full)}}.
Their sizes are  given in \tbl{UAV_datasets}.  They only differ
in the  number of synthetic and  negative samples used for training and testing.
The ratio  of positive to negative  samples in the first  dataset is better
balanced than in  the second one. For \emph{UAV-200  {\scriptsize (small)}}, we
therefore express  our results in  terms of \emph{accuracy}, which  is commonly
used in Domain Adaptation and can be computed as
\begin{equation}
\text{Accuracy} = \frac{\# \text{  correctly classified examples}}{\# \text{ all
    examples}} \; \; .
\label{eq:accuracy}
\end{equation}
Using  this standard  metric  facilitates the  comparison  against the  baseline
methods  whose publicly  available  implementations  only output  classification
accuracy.

In real detection tasks, however, training  datasets are  typically quite
unbalanced, since one usually encounters many negative windows for each positive
example. \emph{UAV-200~{\scriptsize (full)}}  reflects this more realistic scenario,
in which  the accuracy metric is  poorly-suited. For this dataset,  we therefore
compare various approaches in terms of {\it precision-recall}.  Precision corresponds to
 the number of true  positives detected by the  algorithm divided by
the total number  of detections. Recall is the number  of true positives divided
by the number of test examples  labeled as positive. Additionally, we report the
\emph{Average Precision}  (AP), which is  computed as $\int_{0}^{1}  p(r) dr$,
where $p$ and $r$ denote precision and recall, respectively.

For this experiment, we follow the supervised Domain Adaptation scenario. In other
words,  training data  is  available  with labels  for  both  source and  target
domains. 

\begin{table}[!t]
	\centering
	\begin{tabularx}{\linewidth}{XM{0.5cm}M{0.8cm}M{0.9cm}M{0.6cm}M{0.9cm}}
		\toprule
		\multirow{3}{*}{Dataset} & \multicolumn{3}{c}{Training} & \multicolumn{2}{c}{Testing} \\
		\cmidrule(lr){2-4}  
		\cmidrule(lr){5-6}  
		& \multicolumn{2}{c}{Pos} & Neg & Pos & Neg \\
		& \scriptsize{(Real)} & \scriptsize{(Synthetic)} & \scriptsize{(Real)} & \scriptsize{(Real)} & \scriptsize{(Real)} \\
		\midrule
		UAV-200 \scriptsize{(full)} & 200 & 32800 & 190000 & 3100 & 135000 \\
		UAV-200 \scriptsize{(small)} & 200 & 1640 & 9500   & 3100 & 6750   \\
		\bottomrule
	\end{tabularx}
	\caption{{\bf Statistics of our two UAV datasets.} Note that \emph{UAV-200  {\scriptsize (small)}} is more balanced than \emph{UAV-200  {\scriptsize (full)}}. }
	\label{tbl:UAV_datasets}
	\vspace{-0.3cm}
\end{table}

\begin{figure}[!t]
	\centering
	\begin{tabular}{c}
	\vspace{-0.1cm}\hspace{-0.15cm}\includegraphics[width = \linewidth]{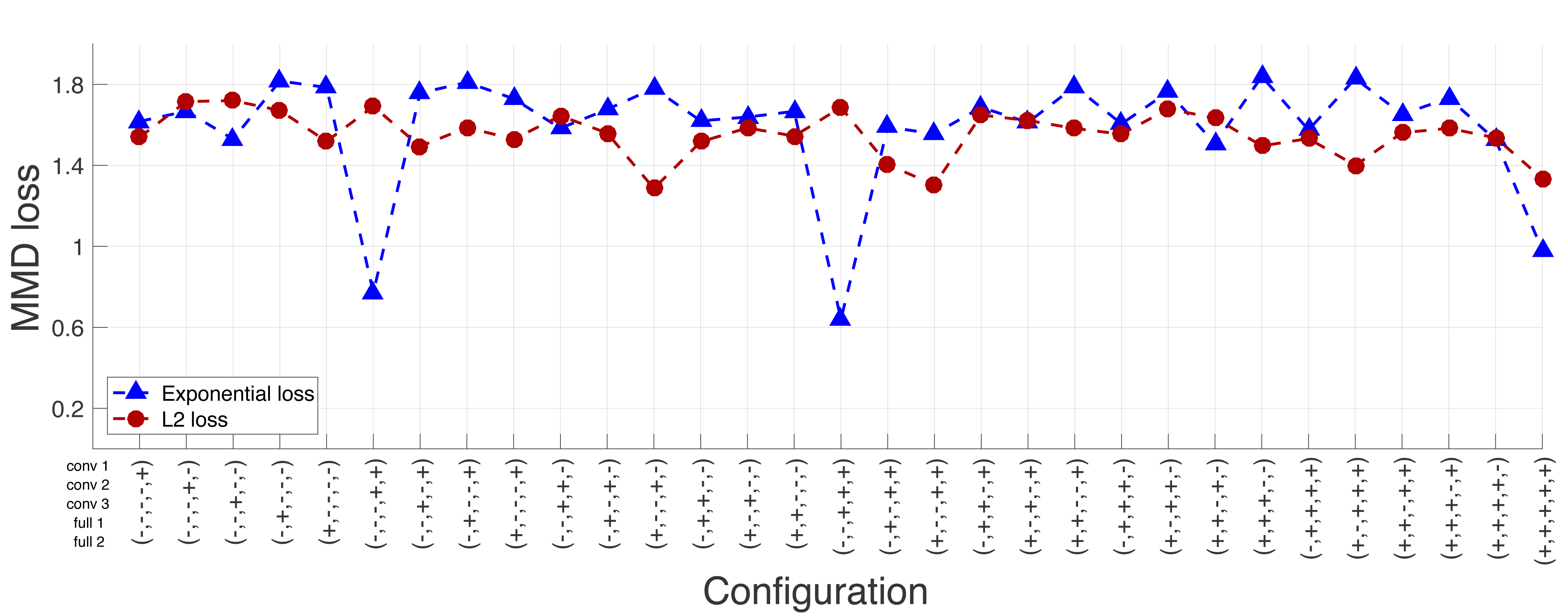} \\
	\hspace{-0.15cm}\includegraphics[width = \linewidth]{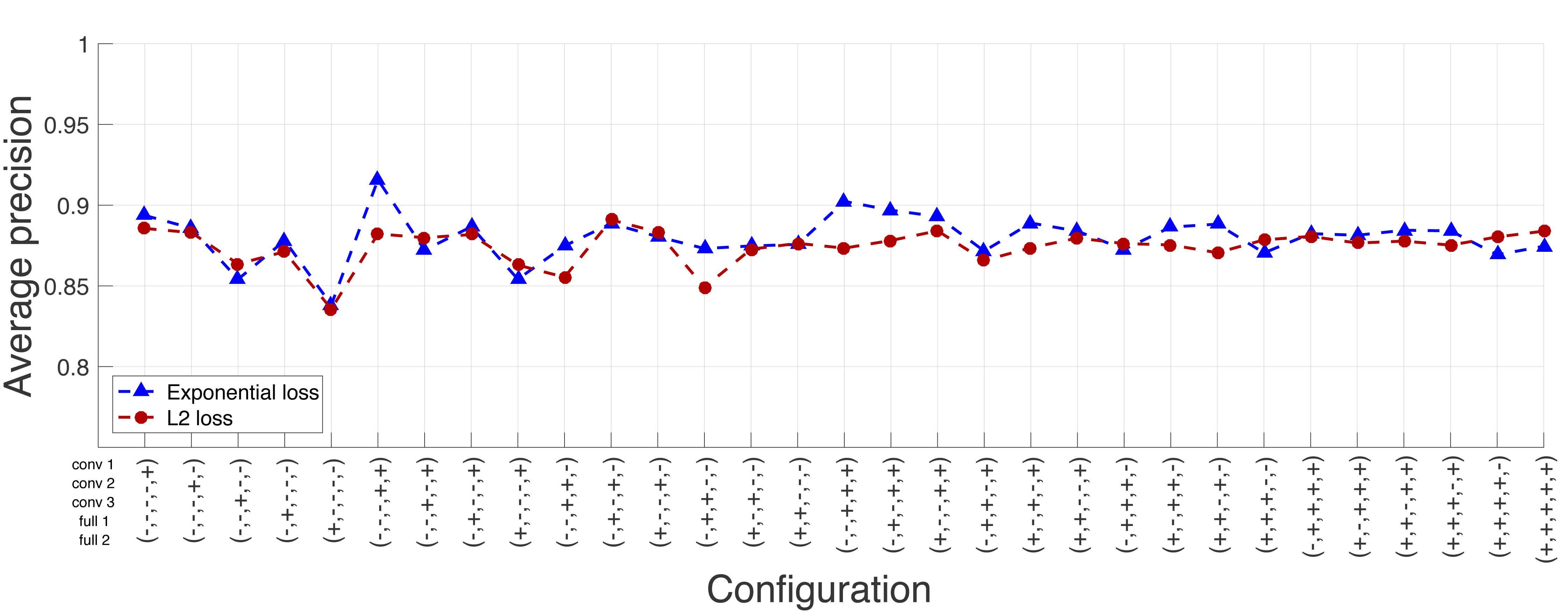} \\
	\end{tabular}
	\caption{{\bf Evaluation of the best network architecture.} \ms{\hl{Top:} The $y$-axis corresponds to the $\mathop{\text{MMD}}^2$ loss between the outputs of the corresponding streams that operate on real and synthetic data, respectively. \hl{Bottom:} Here the $y$-axis corresponds to the AP on validation data ($500$ positive and $1500$ negative examples). Note that low values of MMD tend to coincide with high AP values.} The $x$-axis denotes the network configuration, where a `$+$' sign indicates that the corresponding network layers are regularized with a loss function and a `$-$' sign that the weights are shared for the corresponding layers. (Best seen in color)}
	\label{fig:loss_sel}
	\vspace{-0.3cm}
\end{figure}


\subsubsection{Network Design}
\label{sec:networkDesign}

\comment{Since  drone detection  is  a  relatively new  problem,  there  is no  generally
accepted network architecture, and we had  to design our own.  As illustrated by \fig{streams}}
Our network  consists of two streams, one for  the source data and  one  for  the target data, as illustrated by \fig{streams}.
Each  stream  is  a CNN  that  comprises  three
convolutional and max-pooling layers, followed by two fully-connected ones.  The
classification layer  encodes a hinge  loss, which  was shown to  outperform the
logistic loss in practice for some tasks~\cite{Jin14,Jaderberg15}.

As discussed  above, some  pairs of  layers in  our two-stream  architecture may
share their  weights while  others do  not, and  we must  decide upon  an optimal
arrangement. \comment{In a  real-world case, we would not have  access to validation data
and we  therefore need a way  to do so without  it.} To this end,  we trained one
model  for every  possible combination.  \ms{In every  case, we  implemented our
  regularizer using  either the  $L_2$ loss  of Eq.~\ref{eq:L2_loss_psi}  or the
  exponential  loss   of  Eq.~\ref{eq:exp_loss_psi}.    After training, we  then   computed  the
  $\mathop{\text{MMD}}^2$ value between  the  output of  both  streams for  each
  configuration.} We plot the results in \fig{loss_sel}~\hl{(top)}, with the $+$
and  $-$ signs  indicating whether  the weights  are stream-specific  or shared.
Since we use a common classification  layer, the $\mathop{\text{MMD}}^2$
value  ought to  be small  when  our architecture  accounts well  for the  domain
shift~\cite{Tzeng14}. It therefore makes sense  to choose the configuration that
yields the  smallest $\mathop{\text{MMD}}^2$ value.   In this case,  it happens
when using  the exponential loss to  connect the first three  layers and sharing
the weights of the others. Our intuition is that, even though the synthetic
and real  images feature the same  objects, they differ in  appearance, which is
mostly  encoded by  the first  network layers.   Thus, allowing  the weights  to
differ in these layers yields good  adaptative behavior, as will be demonstrated
in Section~\ref{sec:droneEval}.

\ms{As  a sanity  check, we used validation data ($500$ positive and $1500$ negative examples) to confirm that this MMD-based criterion
 reflects the best architecture choice. In \fig{loss_sel}~\hl{(bottom)},  we plot the real detection accuracy
 as a  function of  the chosen  configuration.  The best
 possible accuracies are 0.916 and 0.757 on the validation and test data, respectively, whereas the ones corresponding to our MMD-based choice
 are 0.902 and 0.732, which corresponds to the second best architecture.  Note that the MMD of the best solution also is very low. Altogether, we believe that this evidences that our MMD-based criterion provides an effective alternative to select the right architecture in the absence of validation data.}
 
\subsubsection{Evaluation}
\label{sec:droneEval}

We  first   compare  our  approach   to  other  Domain  Adaptation   methods  on
\emph{UAV-200 \scriptsize{(small)}}.   As can be seen  in \tbl{Domain_adapt}, it
significantly outperforms many state-of-the-art  baselines in terms of accuracy.
In particular, we believe that  outperforming DDC~\cite{Tzeng14} goes a long way
towards  validating  our hypothesis  that  modeling  the  domain shift  is  more
effective than trying to  be invariant to it.  This is  because, as discussed in
Section~\ref{sec:related}, DDC  relies on  minimizing the  MMD loss  between the
learned  source and  target representations  much as  we do,  but uses  a single
stream  for  both source  and  target  data.  In  other  words,  except for  the
non-shared weights,  it is the method  closest to ours. Note,  however, that the
original DDC paper used a slightly different network architecture than ours.  To
avoid any bias, we therefore modified this architecture so that it matches ours.

\begin{table}[!t]
	\centering
	\begin{tabularx}{\linewidth}{XM{1.5cm}}
		\toprule
		& \multicolumn{1}{c}{Accuracy} \\
		\midrule
		ITML~\cite{Saenko10} 					& 0.60 \\
		ARC-t assymetric~\cite{Kulis11} 		& 0.55 \\
		ARC-t symmetric~\cite{Kulis11} 			& 0.60 \\
		HFA~\cite{Li14b} 						& 0.75 \\
		DDC~\cite{Tzeng14} 						& 0.89 \\
		Ours 									& {\bf 0.92} \\
		\bottomrule
	\end{tabularx}
	\caption{Comparison  to  other  domain   adaptation  techniques  on  the
          \emph{UAV-200 \scriptsize{(small)}} dataset.}
	\label{tbl:Domain_adapt}
	\vspace{-0.2cm}
\end{table}

We then turn to the complete dataset \emph{UAV-200 \scriptsize{(full)}}. In this
case, the  baselines whose  implementations only  output accuracy  values become
less relevant because it is not a  good metric for unbalanced data. We therefore
compare  our approach  against  DDC~\cite{Tzeng14},  which we  found  to be  our
strongest competitor in  the previous experiment, and against  the Deep Learning
approach of~\cite{Rozantsev15b}, which also tackles the drone detection problem.
We also turn  on and off some of  our loss terms to quantify  their influence on
the final performance.  We  give the results in  \tbl{Baselines_summary}.  In short,
all loss  terms contribute  to improving  the AP  of our  approach, which
itself outperforms all the baselines by large margins. More specifically, we get
a 10\% boost over DDC and a 20\%  boost over using real data only.  By contrast,
simply   using   real   and   synthetic   examples   together,   as   was   done
in~\cite{Rozantsev15b}, does not yield significant improvements. \pf{Note that
  dropping the  terms linking  the weights in  corresponding layers  while still
  minimizing the MMD loss ({\emph Loss:} {\small$L_{s} + L_{t} + L_{MMD}$}) performs
worse than using our full loss function. We attribute this to overfitting of the
target stream.}

\begin{table}[!t]
	\centering
	\begin{tabularx}{\linewidth}{XM{1.5cm}}
		\toprule
		& AP \\
		\multicolumn{2}{r}{\scriptsize{(Average Precision)}} \\
		\midrule
		\emph{CNN \scriptsize{(trained on Synthetic only ({\bf S}))}} & 0.314 \\
		\emph{CNN \scriptsize{(trained on Real only ({\bf R}))}} & 0.575 \\
		\emph{CNN \scriptsize{(pre-trained on {\bf S} and fine-tuned on {\bf R})}:} & \\
		\qquad\emph{Loss: } $L_{t}$ \comment{(\eqt{cost_r})} & 0.612 \\
		\qquad\emph{Loss: } $L_{t} + L_{w}$\comment{(\eqts{cost_r}{cost_rf})} \scriptsize{\emph{(with fixed source CNN)}}& 0.655 \\
		\emph{CNN \scriptsize{(pre-trained on {\bf S} and fine-tuned on {\bf R} and {\bf S}:)}} & \\
		\qquad\emph{Loss: } $L_{s} + L_{t}$\comment{(\eqts{cost_rfs}{cost_r})} \cite{Rozantsev15b}	& 0.569 \\
		DDC~\cite{Tzeng14} \scriptsize{\emph{(pre-trained on {\bf S} and fine-tuned on {\bf R} and {\bf S})}} & 0.664 \\
		\midrule	
		\multicolumn{2}{l}{\emph{Our approach \scriptsize{(pre-trained on {\bf S} and fine-tuned on {\bf R} and {\bf S})}}} \\
		\qquad\emph{Loss: } $L_{s} + L_{t} + L_{w}$\comment{(\eqtss{cost_rfs}{cost_r}{cost_rf})}		& 0.673 \\
		\qquad\emph{Loss: } $L_{s} + L_{t} + L_{MMD}$\comment{(\eqtss{cost_rfs}{cost_r}{cost_rfssm})} 	& 0.711 \\
		\qquad\emph{Loss: } $L_{s} + L_{t} + L_{w} + L_{MMD}$\comment{(\eqtsss{cost_rfs}{cost_r}{cost_rf}{cost_rfssm})} & {\bf 0.732} \\ 
		\bottomrule
	\end{tabularx}
	\caption{Comparison of  our  method  against  several baselines  on  the
          \emph{UAV-200   \scriptsize{(full)}}   dataset.    As   discussed   in
          \sect{method}, the terms $L_{s}$, $L_{t}$, $L_{w}$, and $L_{MMD}$
          correspond  to  the   elements  of  the  loss   function,  defined  in
          \eqtsss{cost_rfs}{cost_r}{cost_rf}{cost_rfssm}, respectively.}
	\label{tbl:Baselines_summary}
\end{table}

\comment{For the sake of comparison we also provide the average precision scores for the simple methods, such as conventional CNNs, trained solely on real or synthetic data. Our experiments suggest that adding synthetic data allows to improve the detection accuracy by 20\%, while simply combining real and synthetic examples together does not improve the performance much, which was proposed by~\cite{Rozantsev15b} for smaller amounts of synthetic data. We also provide with the performance of the Deep Domain Confusion algorithm~\cite{Tzeng14}, which differs from our method due to shared weights of $CNN_t$ and $CNN_s$. As we can see from \tbl{Baselines_summary}, allowing weights of the network to change improves performance of the detection algorithm. 

An interesting baseline to our method will be to allow weights of both $\text{CNN}_t$ and $\text{CNN}_s$ to be learned independently, while the outputs of their last fully-connected layers are regularized with an MMD loss (\eqt{cost_rfssm}). \tbl{Baselines_summary} shows that our method is able to outperform this baseline (denoted as `Pair of CNNs with \emph{Loss: }\eqtss{cost_r}{cost_rfs}{cost_rfssm}'). This likely happens due to the over-fitting of $\text{CNN}_t$ to the target data.}

\subsubsection{Influence of the Number of Samples}

\begin{figure}[!t]
	\centering
	\begin{tabular}{cc}
		\hspace{-0.15cm}\includegraphics[width = 0.49\linewidth]{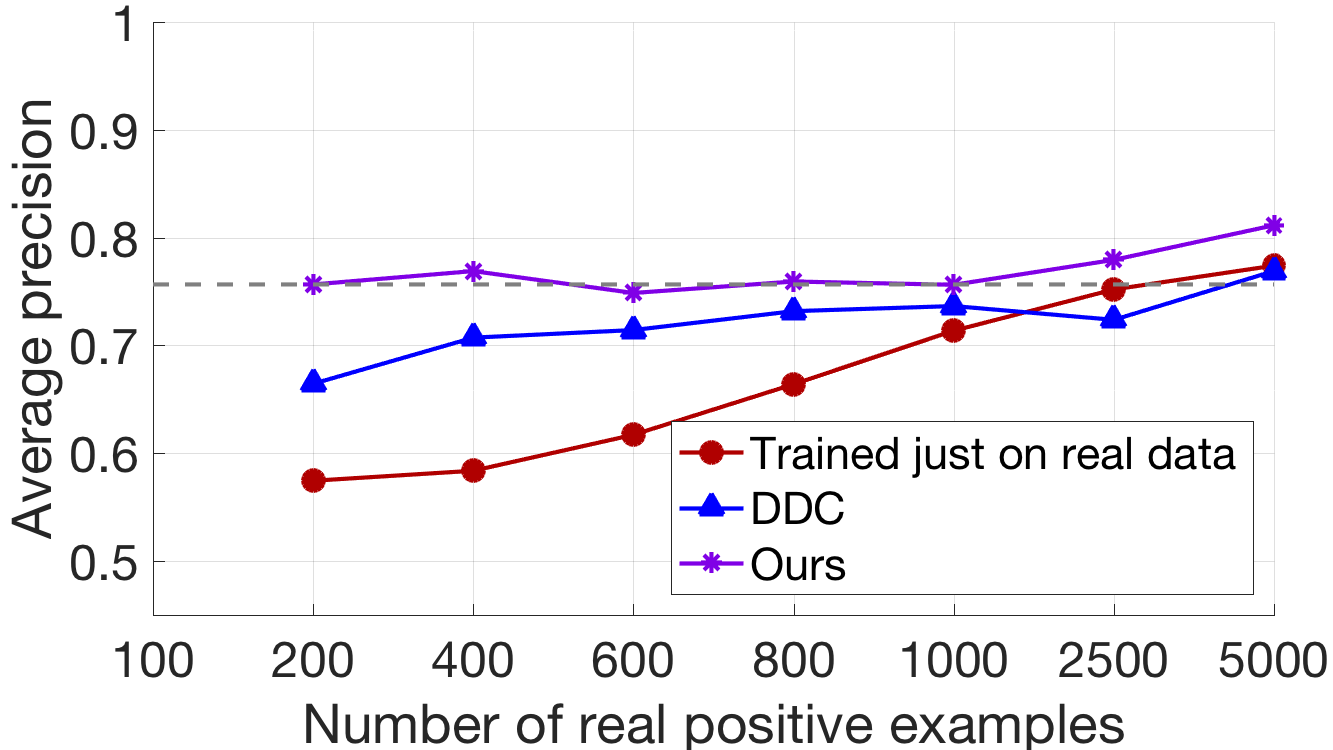} &
		\hspace{-0.3cm}\includegraphics[width = 0.49\linewidth]{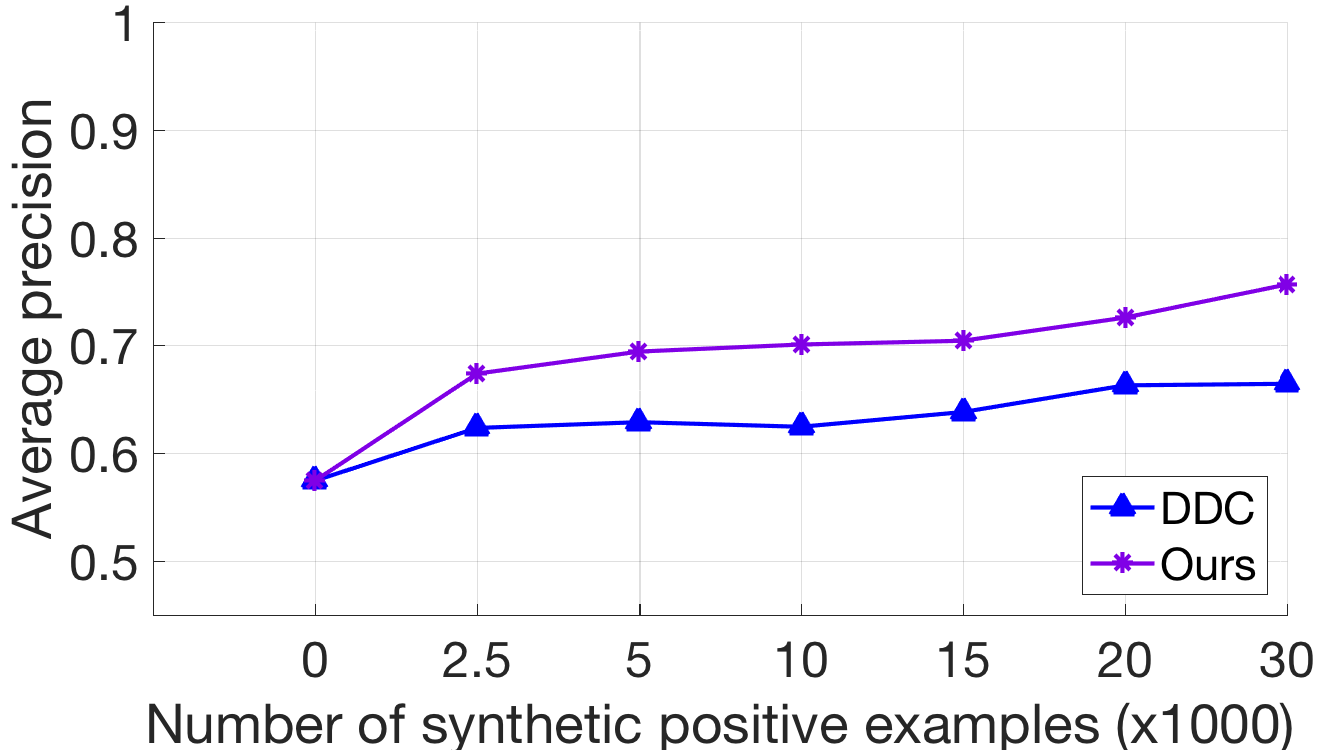} \\
	\end{tabular}
	\caption{{\bf Influence  of the  ratio of  synthetic to  real data.}
            \hl{Left:} AP of our approach  (violet stars), DDC (blue triangles),
            and training using real data only (red circles) as a function of the
            number of real samples used given a constant number of synthetic ones.
            \hl{Right:}  AP  of  our  approach   (violet  stars)  and  DDC  (blue
            triangles) as  a function of  the number of synthetic  examples used
            given a small and constant number of real one. (Best seen in color)}
	\label{fig:comp2real}
	\vspace{-0.3cm}
\end{figure}

Using synthetic data in the UAV detection scenario is motivated by the fact that
it  is hard  and time  consuming  to collect  large  amounts of  real data.   We
therefore evaluate the influence of the ratio of synthetic to real data. To this
end, we first fix the number of synthetic samples to 32800, as in \emph{UAV-200
  \scriptsize{(full)}}, and vary the amount of  real positive samples from 200 to
5000.  The results of this experiment are reported in \fig{comp2real}(left), where
we again compare our approach to DDC~\cite{Tzeng14} and to the same CNN model trained on the real
samples only.  Our  model always outperforms the one trained  on real data only.
This suggests that  it remains capable of leveraging the  synthetic data, even though
more real data  is available, which is  not the case for DDC.  More importantly,
looking  at the  leftmost point  on our  curve shows  that, with  only 200  real
samples, our  approach performs similarly to,  and even slightly better  than, a
\ms{single-stream} model trained using 2500 real samples. In other words, one only needs to collect
5-10\% of labeled training data to obtain good results with our approach, which,
we believe, can have a significant impact in practical applications.

\fig{comp2real}(right) depicts the results of an experiment where we fixed
the number  of real samples  to 200 and increased  the number of  synthetic ones
from 0 to 32800.  Note that the AP of our approach steadily increases as
more synthetic data is used.  DDC also improves, but we systematically outperform
it except when we use no synthetic samples, in which case both approaches reduce
to a single-stream CNN trained on real data only.

\subsection{Unsupervised Domain Adaptation on \it{Office}}

To demonstrate  that our approach extends  to the unsupervised case,  we further
evaluate it on the \emph{Office} dataset,  which is a standard domain adaptation
benchmark for image classification. Following  standard practice, we express our
results in terms of accuracy, as defined in Eq.~\ref{eq:accuracy}.

The  \emph{Office} dataset~\cite{Saenko10}  comprises  three  different sets  of
images (Amazon, DSLR,  Webcam) featuring 31 classes  of objects. \fig{office_im}
depicts  some images  from  the  three different  domains.  For our  experiments, we  used the  ``fully-transductive'' evaluation
protocol  proposed  in~\cite{Saenko10},  which  means using  all  the  available
information on  the source  domain and having  no labels at  all for  the target
domain.

\ms{In addition to the results obtained using our MMD regularizer of \eqt{cost_rfssm}, and for  a  fair  comparison with~\cite{Ganin15},  which  achieves  state-of-the-art
results on this dataset, we also report results obtained by replacing the MMD loss 
with one based on the domain confusion classifier advocated  in~\cite{Ganin15}. We used the same architecture as in~\cite{Ganin15} for this classifier.}

\begin{figure}[!t]
	\centering
	\begin{tabular}{cc}
		\hspace{-0.15cm}\includegraphics[width = 0.44\linewidth]{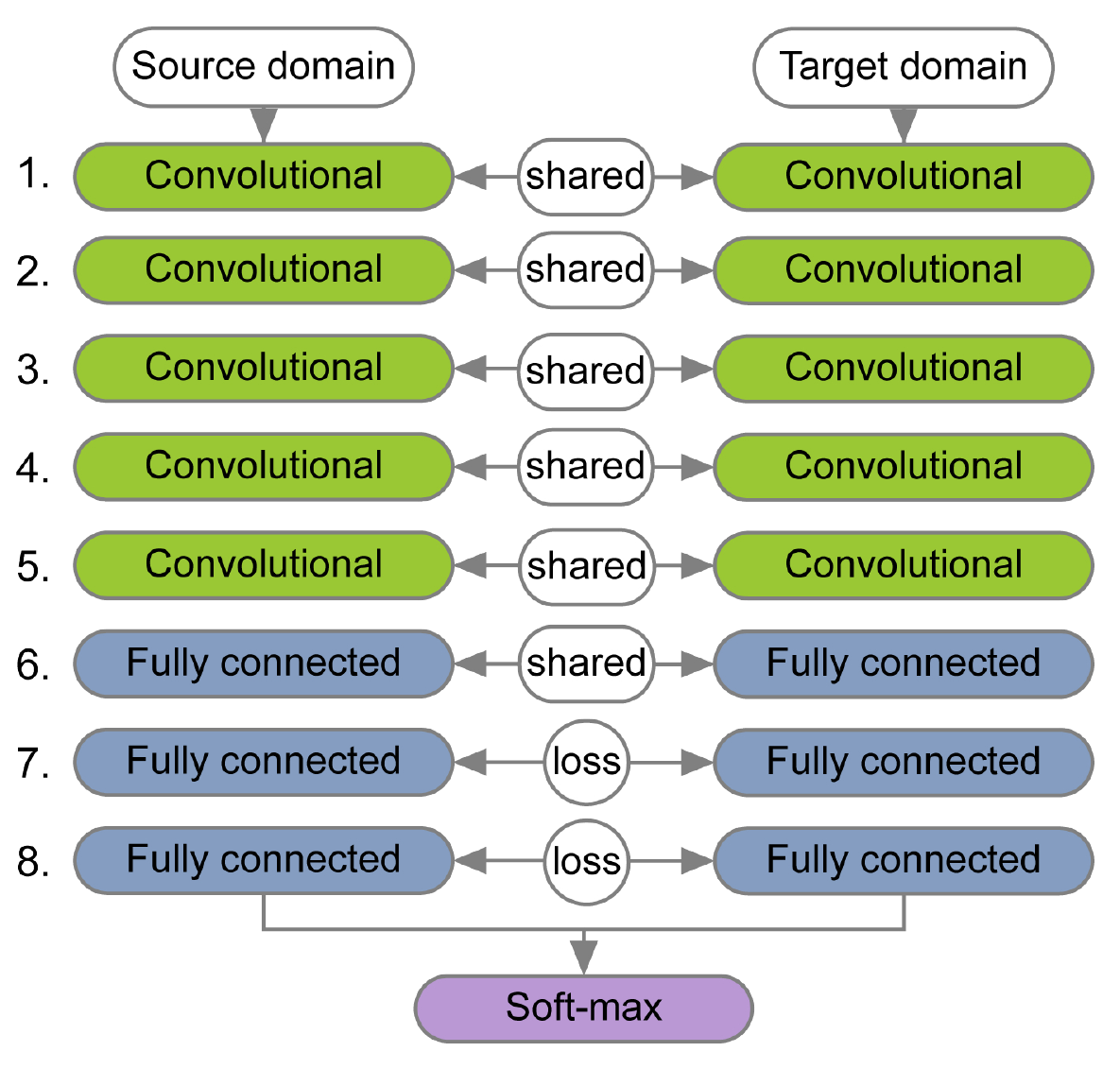} & 
		\hspace{-0.30cm}\includegraphics[width = 0.56\linewidth]{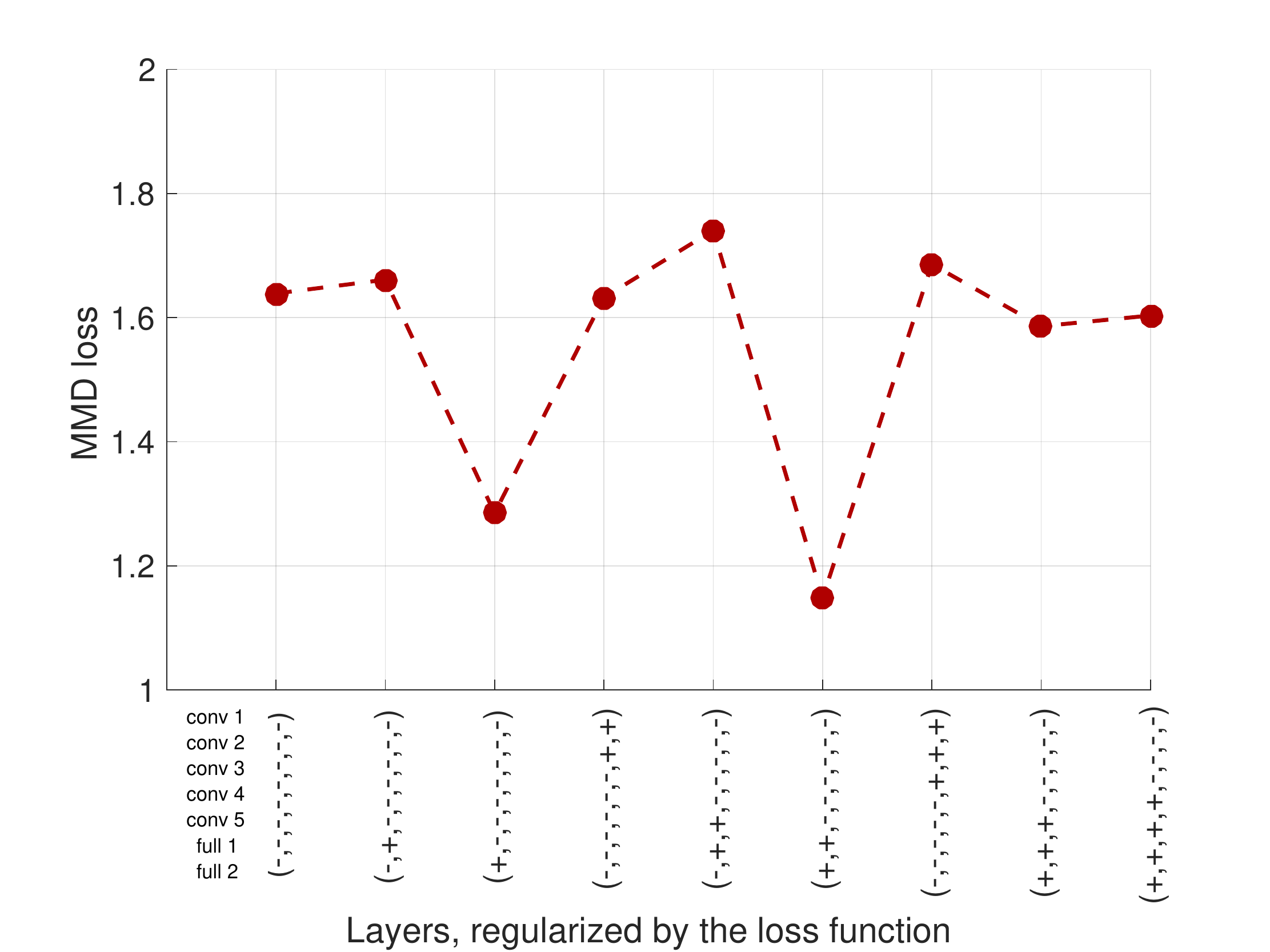} \\
		(a) & (b) \\
	\end{tabular}
	\caption{Office dataset.  \hl{(a)} The  network architecture that proved
          to be the best according \ms{to our MMD-based criterion.} \hl{(b)}
          The  $y$-axis  corresponds  to  the  $\mathop{\text{MMD}}^2$  loss
            between the  outputs of  the corresponding  streams that  operate on
            {\it Amazon} and {\it Webcam}, respectively.  The $x$-axis  
          describes the configuration, as in \fig{loss_sel}.}
	\label{fig:office_selection}
	\vspace{-0.3cm}
\end{figure}

\fig{office_selection}(a) illustrates the network  architecture we used for this
experiment.    \ms{ Each    stream    corresponds   to    the    standard    AlexNet
CNN~\cite{Krizhevsky12}. As in~\cite{Tzeng14,Ganin15}, we start with  the model pre-trained on ImageNet and fine tune it.}  However, instead of forcing the weights of both streams to  be shared,  we allow  them  to deviate  as discussed  in \sect{method}.   To identify which layers should share their weights and which ones should {\it not}, we used  the MMD-based criterion introduced  in Section~\ref{sec:networkDesign}. In  \fig{office_selection}(b), we  plot  the $\mathop{\text{MMD}}^2$  value as  a function  of  the    configuration  on  the  \emph{Amazon}  $\rightarrow$ \emph{Webcam} scenario, as we did for the drones in \fig{loss_sel}. In this case, not sharing   the   last   two    fully-connected   layers   achieves   the   lowest $\mathop{\text{MMD}}^2$  value, and  this  is  the configuration  we  use for  our experiments on this dataset.

\newcommand{\offwidth}{1.00cm}

\begin{figure}[!t]
	\centering
	\begin{tabular}{c}
		\begin{tabular}{cccccccc}
			\hspace{-0.30cm}\rotatebox{90}{\scriptsize{Amazon}} & 
			\hspace{-0.15cm}\includegraphics[width = \offwidth]{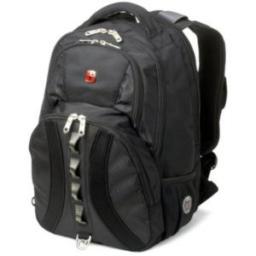} &
			\hspace{-0.30cm}\includegraphics[width = \offwidth]{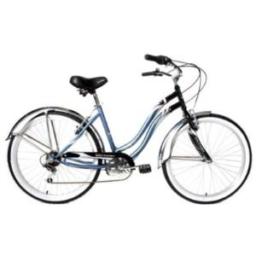} &
			\hspace{-0.30cm}\includegraphics[width = \offwidth]{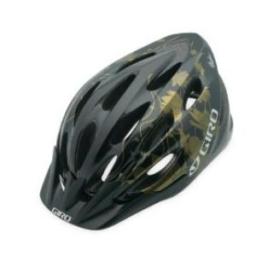} &
			\hspace{-0.30cm}\includegraphics[width = \offwidth]{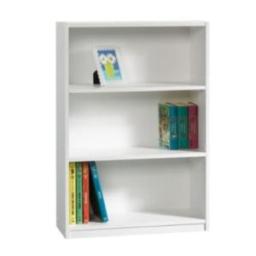} &
			\hspace{-0.30cm}\includegraphics[width = \offwidth]{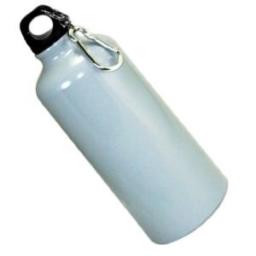} &
			\hspace{-0.30cm}\includegraphics[width = \offwidth]{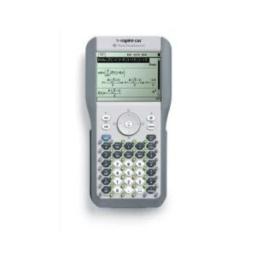} &
			\hspace{-0.30cm}\includegraphics[width = \offwidth]{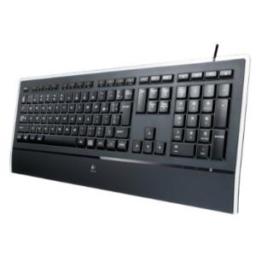} \\
			\hspace{-0.30cm}\rotatebox{90}{\scriptsize{Webcam}} & 
			\hspace{-0.15cm}\includegraphics[width = \offwidth]{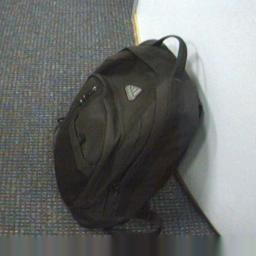} &
			\hspace{-0.30cm}\includegraphics[width = \offwidth]{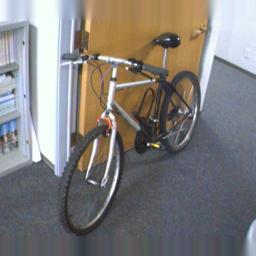} &
			\hspace{-0.30cm}\includegraphics[width = \offwidth]{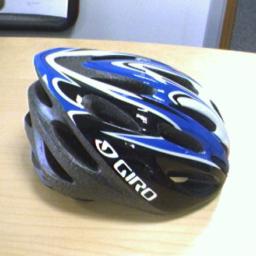} &
			\hspace{-0.30cm}\includegraphics[width = \offwidth]{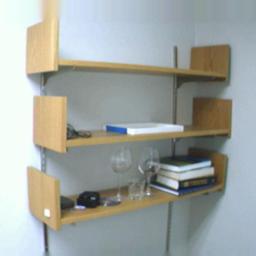} &
			\hspace{-0.30cm}\includegraphics[width = \offwidth]{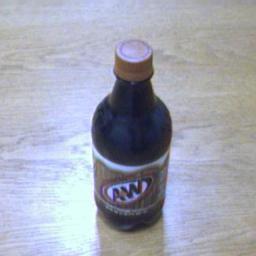} &
			\hspace{-0.30cm}\includegraphics[width = \offwidth]{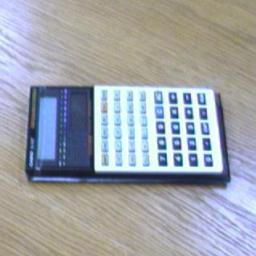} &
			\hspace{-0.30cm}\includegraphics[width = \offwidth]{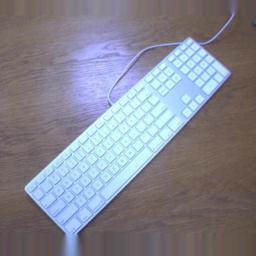} \\
			\hspace{-0.30cm}\rotatebox{90}{\scriptsize{DSLR}} & 
			\hspace{-0.15cm}\includegraphics[width = \offwidth]{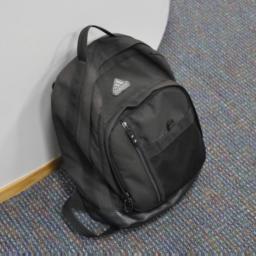} &
			\hspace{-0.30cm}\includegraphics[width = \offwidth]{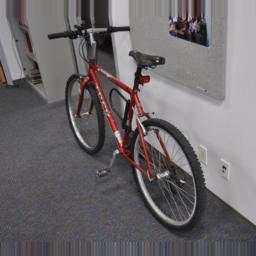} &
			\hspace{-0.30cm}\includegraphics[width = \offwidth]{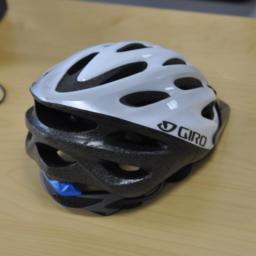} &
			\hspace{-0.30cm}\includegraphics[width = \offwidth]{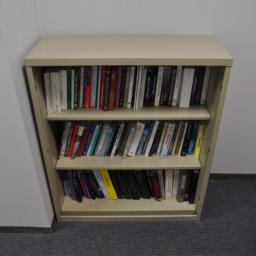} &
			\hspace{-0.30cm}\includegraphics[width = \offwidth]{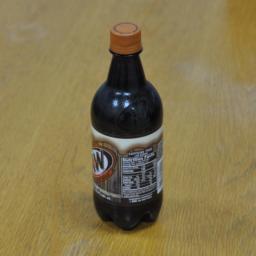} &
			\hspace{-0.30cm}\includegraphics[width = \offwidth]{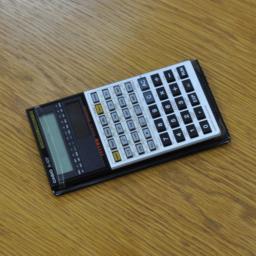} &
			\hspace{-0.30cm}\includegraphics[width = \offwidth]{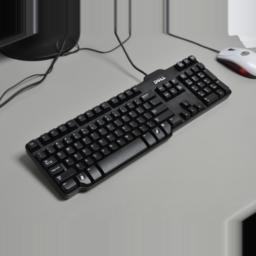} \\
		\end{tabular} \\
	\end{tabular}
	\caption{Some examples from three domains in the Office dataset.
		}
	\label{fig:office_im}
\end{figure}

In  \tbl{unsup_OF}, we  compare  our approach  against  other Domain  Adaptation
techniques on  the three  commonly-reported source/target pairs.  It outperforms
them on all the pairs.  More importantly, the comparison against GRL~\cite{Ganin15} confirms that allowing the weights not to be shared increases accuracy.

\begin{table}[!t]
	\centering
	\begin{tabularx}{\linewidth}{XM{1.1cm}M{1.1cm}M{1.1cm}M{1.1cm}}
		\toprule
		& \multicolumn{4}{c}{Accuracy} \\
		\cmidrule{2-5}
		
		& A $\rightarrow$ W & D $\rightarrow$ W & W $\rightarrow$ D  & Average\\
		GFK~\cite{Gong12b}		  	& 0.214 & 0.691 & 0.650 & 0.518 \\
		DLID~\cite{Chopra13} 		& 0.519 & 0.782 & 0.899 & 0.733 \\
		DDC~\cite{Tzeng14}			& 0.605 & 0.948 & 0.985 & 0.846 \\
		DAN~\cite{Long15b}			& 0.645 & 0.952 & 0.986 & 0.861 \\
		DRCN~\cite{Ghifary16}		& 0.687 & 0.964 & 0.990 & 0.880 \\
		GRL~\cite{Ganin15}			& 0.730 & 0.964 & 0.992 & 0.895 \\
		\cmidrule{2-5}
		Ours \scriptsize{(+ DDC)}	& 0.630 & 0.961 & 0.992 & 0.861 \\
		Ours \scriptsize{(+ GRL)}	& {\bf 0.760} & {\bf 0.967} & {\bf 0.996} & {\bf 0.908} \\
		\bottomrule
	\end{tabularx}
	\caption{Comparison against  other domain  adaptation techniques  on the
          Office benchmark. We evaluate on all 31 categories, following
           the ``fully-transductive'' evaluation protocol~\cite{Saenko10}.}
	\label{tbl:unsup_OF}
	\vspace{-0.3cm}
\end{table}

\comment{
\ms{Additional experiments on the supervised scenario for \emph{Office}, as well
  as  on \emph{MNIST}-\emph{USPS}  for digit  recognition  can be  found in  the
  supplementary material.}}

\subsection{Domain Adaptation on \it{MNIST-USPS}}

The \emph{MNIST}~\cite{LeCun98}  and \emph{USPS}~\cite{Hull94}  datasets for  digit classification
both feature 10 different classes of images corresponding to the 10 digits. They have recently been employed for the task of Domain Adaptation~\cite{Fernando15}.

For this experiment, we used the evaluation protocol of~\cite{Fernando15}, which
involves randomly selecting of $2000$ images from \emph{MNIST} and $1800$ images
from \emph{USPS} and using them interchangeably as source and target domains. As
in~\cite{Fernando15}, we work  in the unsupervised setting, and  thus ignore the
target  domain  labels at  training  time.   
Following~\cite{Long13a}, as  the image patches  in the \emph{USPS}  dataset are
only $16\times16$ pixels,  we rescaled the images from \emph{MNIST}  to the same
size and applied $L_2$ normalization of the pixel intensities.  
\ms{For this experiment, we relied on the standard CNN architecture of~\cite{LeCun98} and
employed our MMD-based criterion to determine which layers  should not
share their weights. We found that allowing all layers of the network not to share their weights yielded the best performance.
}

\comment{
For our
tests, we rely on the  standard CNN architecture of~\cite{LeCun98} and
followed a  similar validation  procedure to that used for the previous two datasets to determine  which layers  should not
share their weights. Specifically, as for the \emph{Office} dataset, since there is no explicit validation set, we randomly sampled $5$ images for each class from the target domain and used them to select the best network architecture. We found that allowing all layers of the network not to share their weights yielded the best performance. \fig{m_vs_u}(a) depicts the final structure of the CNN.  In \fig{m_vs_u}(b), we provide the results of some of the validation experiments.

\begin{figure}[!t]
	\centering
	\begin{tabular}{cc}
		\hspace{-0.15cm}\includegraphics[width = 0.54\linewidth]{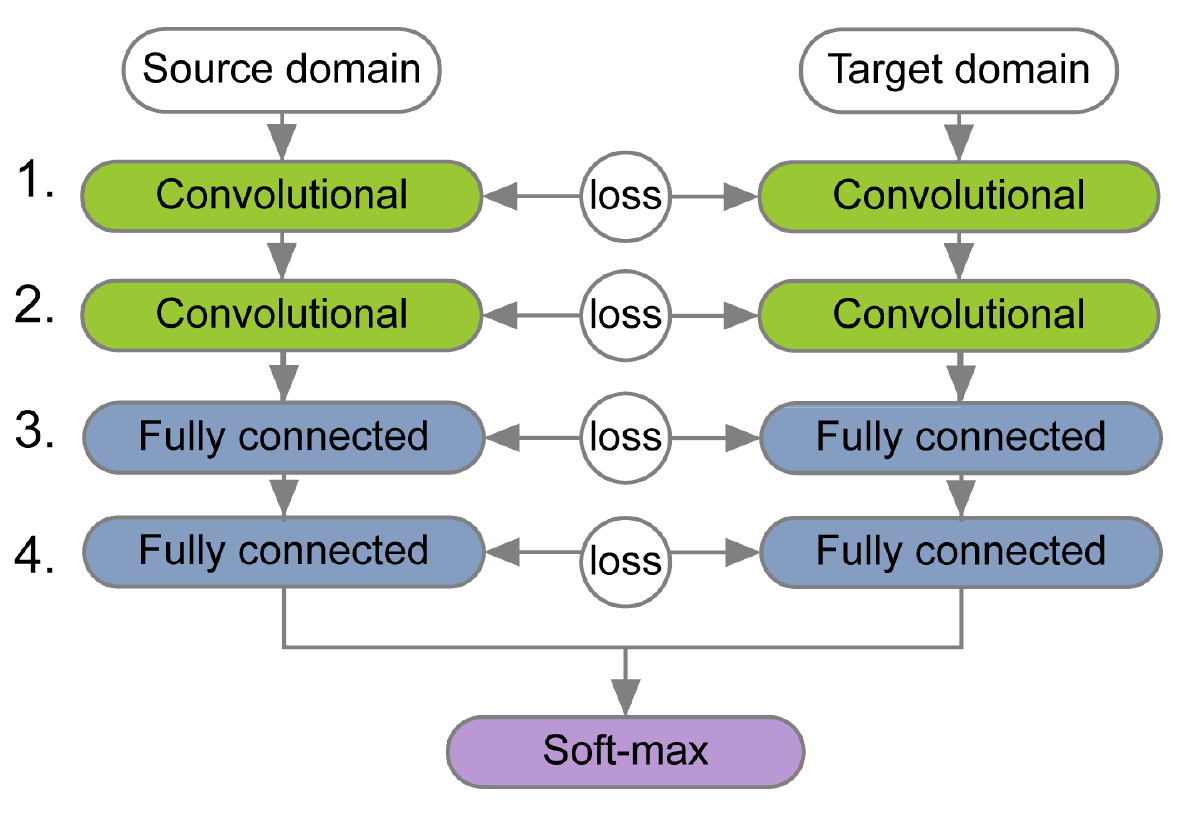} & 
		\hspace{-0.30cm}\includegraphics[width = 0.44\linewidth]{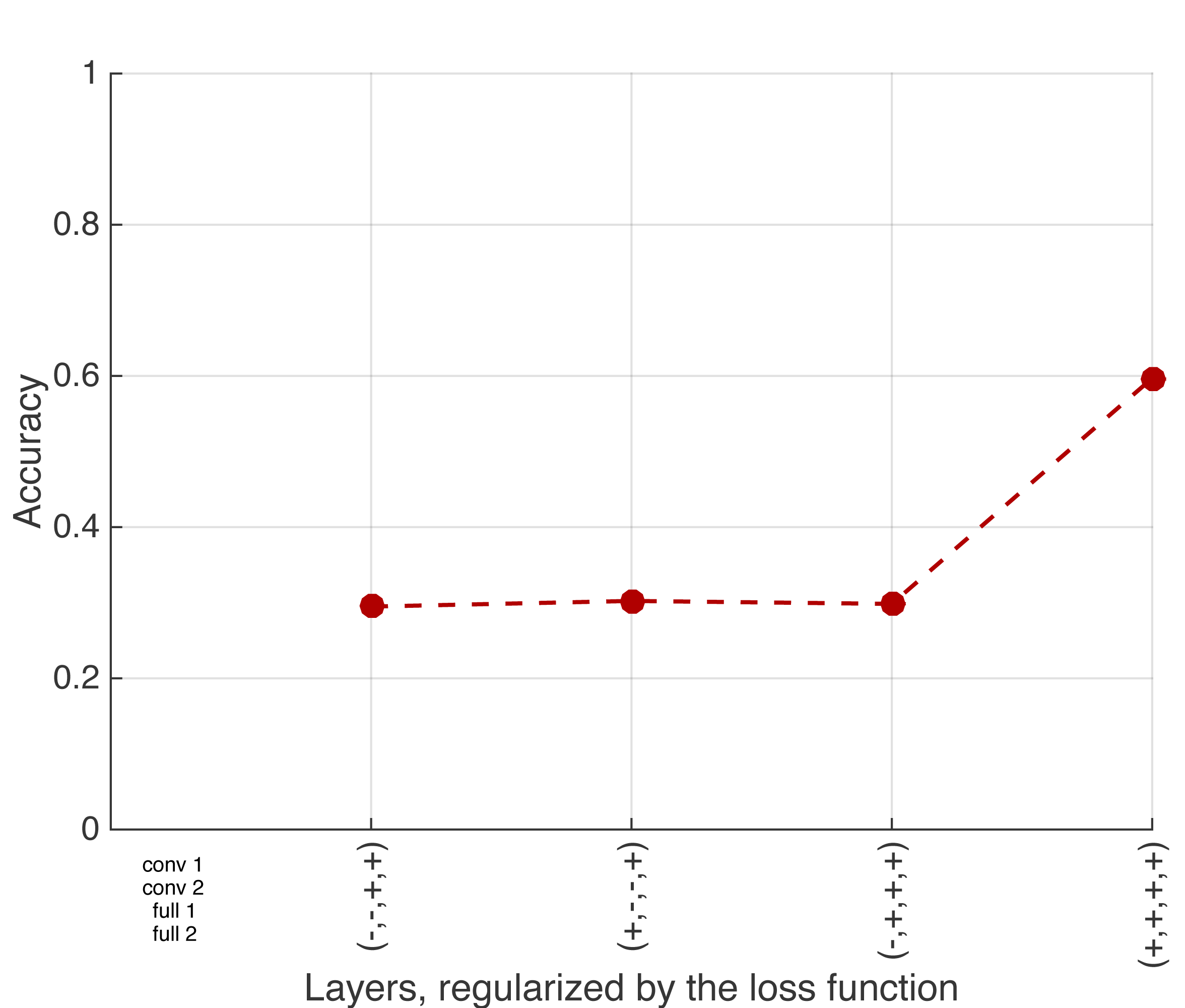} \\
		(a) & (b) \\
	\end{tabular}
	\caption{{\bf Architecture for the MNIST-USPS  dataset.}  \hl{(a)}  Best CNN  architecture based on our validation procedure.
			\hl{(b)}  Results of some of our validation experiments to determine,
			which layers should not share weights  and instead be connected by a
			regularization loss. The $x$-axis denotes the network configuration, where `+' denotes that the weights of a particular layer are \emph{not} shared, while `-' means that they are.}
	\label{fig:m_vs_u}
\end{figure}
}

\ms{In  \tbl{Domain_adapt_MU}, we  compare  our  approach with DDC~\cite{Tzeng14} and with methods that do not rely on deep
networks~\cite{Pan09,Gong12b,Fernando13,Fernando15}. Our method yields superior performance in all cases, which we believe to be due
to its  ability to adapt  the feature representation  to each domain,  while still
keeping these representations close to each other.}

\begin{table}[!t]
	\centering
	\begin{tabularx}{\linewidth}{Xccc}
		\toprule
		& \multicolumn{3}{c}{Accuracy} \\
		\cmidrule{2-4}
		\scriptsize{method} 	& M$\rightarrow$U & U$\rightarrow$M & AVG.\\
		PCA   					& 0.451 & 0.334 & 0.392 \\
		SA~\cite{Fernando13}   	& 0.486 & 0.222 & 0.354 \\
		GFK~\cite{Gong12b}   	& 0.346 & 0.226 & 0.286 \\
		TCA~\cite{Pan09}   		& 0.408 & 0.274 & 0.341 \\
		SSTCA~\cite{Pan09} 		& 0.406 & 0.222 & 0.314 \\
		TSL~\cite{Si10}   		& 0.435 & 0.341 & 0.388 \\
		JCSL~\cite{Fernando15} 	& 0.467 & 0.355 & 0.411 \\
		DDC~\cite{Tzeng14} 		& 0.478 & 0.631 & 0.554 \\
		\cmidrule{2-4}
		Ours 					& {\bf 0.607} & {\bf 0.673} & {\bf 0.640} \\
		\bottomrule
	\end{tabularx}
	\caption{Comparison against  other domain  adaptation techniques  on the
          MNIST+USPS standard benchmark.}
	\label{tbl:Domain_adapt_MU}
	\vspace{-0.1cm}
\end{table}



\subsection{Supervised Facial Pose Estimation}

\begin{figure}[t!]
	\centering
	\begin{tabular}{ccccccc}
		\toprule
		\multicolumn{7}{c}{Synthetic (source domain)} \\
						\includegraphics[width=0.13\linewidth]{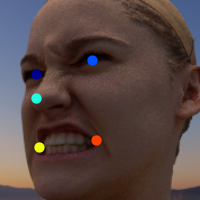} &
		\hspace{-0.35cm}\includegraphics[width=0.13\linewidth]{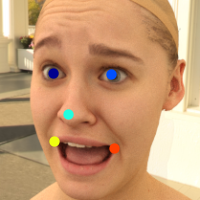} &
		\hspace{-0.35cm}\includegraphics[width=0.13\linewidth]{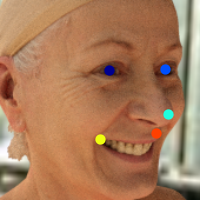} &
		\hspace{-0.35cm}\includegraphics[width=0.13\linewidth]{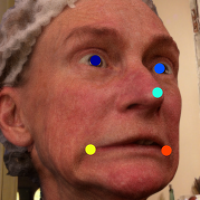} &
		\hspace{-0.35cm}\includegraphics[width=0.13\linewidth]{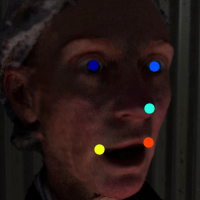} &
		\hspace{-0.35cm}\includegraphics[width=0.13\linewidth]{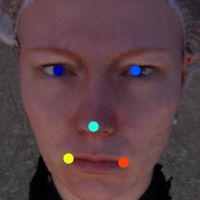} &
		\hspace{-0.35cm}\includegraphics[width=0.13\linewidth]{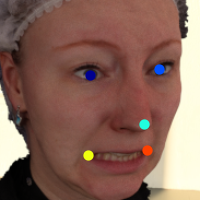} \\
						\includegraphics[width=0.13\linewidth]{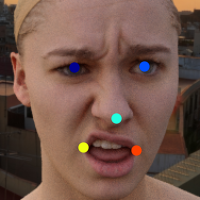} &
		\hspace{-0.35cm}\includegraphics[width=0.13\linewidth]{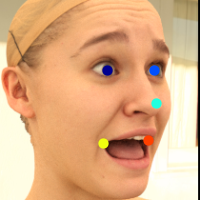} &
		\hspace{-0.35cm}\includegraphics[width=0.13\linewidth]{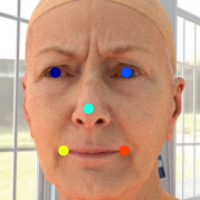} &
		\hspace{-0.35cm}\includegraphics[width=0.13\linewidth]{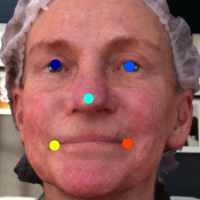} &
		\hspace{-0.35cm}\includegraphics[width=0.13\linewidth]{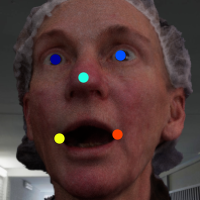} &
		\hspace{-0.35cm}\includegraphics[width=0.13\linewidth]{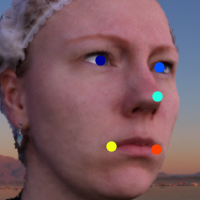} &
		\hspace{-0.35cm}\includegraphics[width=0.13\linewidth]{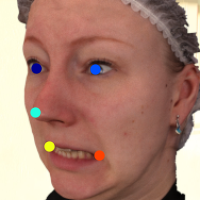} \\
		\midrule
		\multicolumn{7}{c}{Real (target domain)} \\
						\includegraphics[width=0.13\linewidth]{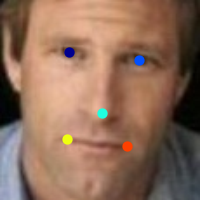} &
		\hspace{-0.35cm}\includegraphics[width=0.13\linewidth]{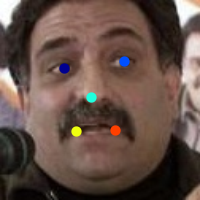} &
		\hspace{-0.35cm}\includegraphics[width=0.13\linewidth]{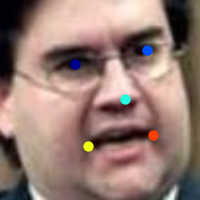} &
		\hspace{-0.35cm}\includegraphics[width=0.13\linewidth]{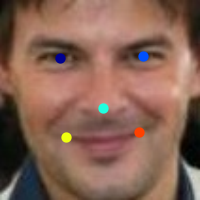} &
		\hspace{-0.35cm}\includegraphics[width=0.13\linewidth]{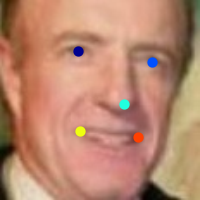} &
		\hspace{-0.35cm}\includegraphics[width=0.13\linewidth]{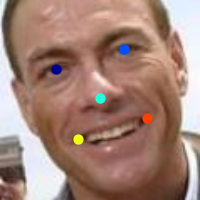} &
		\hspace{-0.35cm}\includegraphics[width=0.13\linewidth]{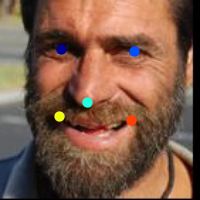} \\
						\includegraphics[width=0.13\linewidth]{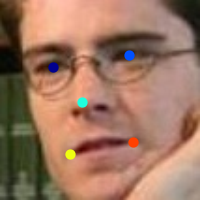} &
		\hspace{-0.35cm}\includegraphics[width=0.13\linewidth]{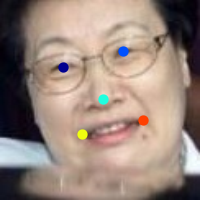} &
		\hspace{-0.35cm}\includegraphics[width=0.13\linewidth]{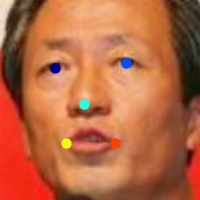} &
		\hspace{-0.35cm}\includegraphics[width=0.13\linewidth]{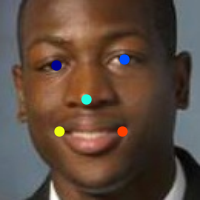} &
		\hspace{-0.35cm}\includegraphics[width=0.13\linewidth]{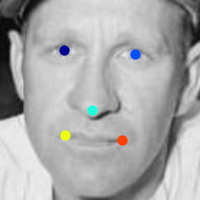} &
		\hspace{-0.35cm}\includegraphics[width=0.13\linewidth]{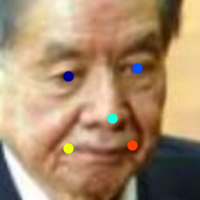} &
		\hspace{-0.35cm}\includegraphics[width=0.13\linewidth]{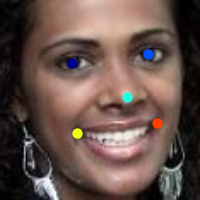} \\
		\bottomrule
	\end{tabular}
	\caption{Samples images from \emph{Source} and \emph{Target} datasets with synthetic and real images respectively.}
	\label{fig:nviso_images}
	\vspace{-0.4cm}
\end{figure}

\ms{To demonstrate  that our method can  be used not only  for classification or
  detection tasks but also for regression ones, we further evaluate it for pose
  estimation purposes. More  specifically, the task we address consists of predicting the
  location of  $5$ facial landmarks given  $50 \times 50$ image patches, such as
  those of~\fig{nviso_images}. 
  \comment{ which can then be  used to compute the 6D pose of
  the camera  with respect to the  faces.}
  To this  end, we train a  regressor to
  predict a 10D vector with two floating point coordinates for each landmark. As
  we did for drones, we use \emph{synthetic} images, such as the ones shown in the top
  portion of~\fig{nviso_images}, as our  source domain and \emph{real}  ones, such as those  shown at the
  bottom,  as  our target  domain.   Both  datasets contain  $\sim10k$  annotated
  images.  We  use all the synthetic samples but only $100$ of the real ones for
  training, and  the remainder for testing.  For more  detail on these
  two datasets,  we refer  the interested reader  to the  supplementary material
  where we also describe the architecture of the regressor we use.}

\pf{In \tbl{nviso_res},  we compare  our Domain Adaptation  results to  those of
  DDC~\cite{Tzeng14}  in terms  of percentage  of correctly  estimated landmarks
  (PCP-score).  Each landmark is considered  to be correctly estimated if it
  is found within a $2$ pixel radius from the ground-truth.}
\ms{Note that, again, by not sharing the weights, our approach outperforms DDC.}

\begin{table}[t!]
	\centering
	\begin{tabularx}{\linewidth}{Xccc}
		\toprule
		& Synthetic & DDC~\cite{Tzeng14} & Ours \\
		\midrule
		Right eye 			& 64.2 & 68.0 & {\bf 71.8} \\
		Left eye 			& 39.3 & 56.2 & {\bf 60.3} \\
		Nose 				& 56.3 & 64.1 & {\bf 64.5} \\
		Right mouth corner 	& 47.8 & 57.6 & {\bf 59.8} \\
		Left mouth corner 	& 42.3 & 55.5 & {\bf 57.7} \\
		\midrule
		Average 			& 50.0 & 60.3 & {\bf 62.8} \\
		\bottomrule
	\end{tabularx}
	\caption{Regression results on facial pose estimation.}
	\label{tbl:nviso_res}
	\vspace{-0.4cm}
\end{table}

\subsection{Discussion}

In all  the experiments  reported above,  allowing the weights  {\it not}  to be
shared in some fraction of  the layers of  our two-stream  architecture boosts
performance.  This validates our initial hypothesis that explicitly modeling the
domain shift is generally beneficial.

However, the  optimal choice  of which  layers should or  should not  share their
weights is application  dependent. In the UAV case, allowing  the weights in the
first two layers to be different  yields top performance, which we understand to
mean that the domain shift is caused  by low-level changes that are best handled
in the early  layers. By contrast, for  the {\it Office} dataset, it  is best to
only  allow  the  weights in  the  last  two  layers  to differ.   This  network
configuration was determined using  \emph{Amazon} and \emph{Webcam} images, such
as those shown  in Fig.~\ref{fig:office_im}.  Close examination  of these images
reveals  that the  differences  between them  are not  simply  due to  low-level
phenomena, such  as illumination  changes, but to  more complex  variations.  It
therefore seems reasonable  that the higher layers of the  network, which encode
higher-level information, should be domain-specific.

Fortunately, we have shown that the  MMD provides us with an effective criterion
to choose the right configuration. This makes our two-steam approach practical, 
even when no validation data is available.


\section{Conclusion}

\vspace{-0.2cm}

In  this paper,  we  have postulated  that Deep  Learning  approaches to  Domain
Adaptation  should not  focus on  learning features  that are  invariant to  the
domain shift, which makes them less discriminative. Instead, we should explicitly
model  the domain  shift. To  prove this,  we have  introduced a  two-stream CNN
architecture, where  the weights of  the streams may or  may not be  shared.  To
nonetheless encode  the fact that both  streams should be related,  we encourage
the non-shared weights  to remain close to being linear  transformations of each
other by introducing an additional loss term.

Our  experiments   on  very   diverse  datasets   have  clearly   validated  our
hypothesis. Our approach consistently yields  higher accuracy than networks that
share all  weights for the source  and target data, both  for classification and
regression.   In  the  future,  we  intend  to  study  if  more  complex  weight
transformations could  help us  further improve our  results, with  a particular
focus  on   designing  effective  constraints   for  the  parameters   of  these
transformations.


{\small
\bibliographystyle{ieee}
\bibliography{string,vision,learning}
}

\end{document}